\documentclass[lettersize,journal]{IEEEtran}
\usepackage{amsmath,amsfonts,amssymb}
\usepackage{algorithmic}
\usepackage{algorithm}
\usepackage{array}
\usepackage[caption=false,font=normalsize,labelfont=sf,textfont=sf]{subfig}
\usepackage{textcomp}
\usepackage{stfloats}
\usepackage{url}
\usepackage{verbatim}
\usepackage{graphicx}
\graphicspath{{./}}
\usepackage[nocompress]{cite}
\usepackage{enumitem}
\usepackage{booktabs}
\usepackage{multirow}
\usepackage{xcolor}
\usepackage{colortbl}
\usepackage{tabularx}
\newcolumntype{C}[1]{>{\centering\arraybackslash}p{#1}}%
\newcolumntype{Y}{>{\centering\arraybackslash}X}%
\definecolor{oursgray}{HTML}{FFF3D1}

\newcommand{\ua}{$\uparrow$}
\newcommand{\da}{$\downarrow$}
\begin{document}

\title{Learning to Assess Heartbeat Observability for mmWave Heart-Rate Sensing}

\author{Yuxuan~Hu, Shilin~Shan, Jianfei~Yang, and Feng~Xu%
\thanks{This work was supported by the Fundamental and Interdisciplinary Disciplines Breakthrough Plan of the Ministry of Education of China under Grant JYB2025XDXM110.}%
\thanks{Yuxuan Hu and Feng Xu are with the Key Laboratory for Information Science of Electromagnetic Waves, Ministry of Education, School of Information Science and Technology, Fudan University, Shanghai 200433, China (e-mail:huyx23@m.fudan.edu.cn; fengxu@fudan.edu.cn). Yuxuan Hu is also with the School of Mechanical and Aerospace Engineering, Nanyang Technological University, Singapore 639798.}%
\thanks{Shilin Shan and Jianfei Yang are with the School of Mechanical and Aerospace Engineering, Nanyang Technological University, Singapore 639798 (e-mail: shilin.shan@ntu.edu.sg; jianfei.yang@ntu.edu.sg).}%
\thanks{\textit{(Corresponding author: Feng Xu, Jianfei Yang.)}}}

\markboth{}%
{Hu \MakeLowercase{\textit{et al.}}: Learning to Assess Heartbeat Observability for mmWave Heart-Rate Sensing}

\maketitle

\begin{abstract}
Contactless heart-rate sensing with millimeter-wave (mmWave) radar requires assessing whether individual measurements support reliable estimation. We study learning to assess heartbeat observability, defined as the readability of the heartbeat component in an acquired phase spectrum, for selective heart-rate estimation. Coherent superposition of scatterer returns can suppress this component even under similar macroscopic observation geometry, motivating assessment directly from acquired measurements. To obtain training supervision across different observability conditions, we develop a controllable multi-scatterer frequency-modulated continuous-wave (FMCW) simulator. Agreement between the dominant heartbeat-band peak and the known heart rate provides an automatic observability label for each simulated measurement. We propose HEAR (Heartbeat Estimation with Assessed Reliability), a compact dual-task Transformer that jointly predicts an observability score and heart rate. Its input combines spectral magnitudes with frequencies relative to the respiration fundamental, providing context for respiratory harmonics. Trained solely on simulated observations, HEAR transfers zero-shot to two public real-world datasets collected at 60 and 120 GHz from 134 subjects. The same learned score supports selective prediction with both HEAR's own heart-rate head and multiple existing estimators. On the 120 GHz dataset, score-based selection reduces the HR head's mean absolute error from 17.9 BPM at full coverage to 1.6 BPM at 50\% coverage. The complete pipeline achieves an end-to-end processing latency of 50.8 ms on an edge device. Project page: \url{https://yuxuanhu9.github.io/HEAR/}.

\end{abstract}

\begin{IEEEkeywords}
Heartbeat observability assessment, selective prediction, multi-task learning, sim-to-real transfer, millimeter-wave radar, contactless heart-rate estimation
\end{IEEEkeywords}

\section{Introduction}

\IEEEPARstart{C}{ontactless} continuous heart-rate monitoring is increasingly important for everyday healthcare~\cite{Mercuri2019}. {Healthcare and companion robots~{\cite{ScienceRobotics001,technologies9010008}} could provide this capability for pediatric and elderly care and chronic-condition management~{\cite{Hu2025_ActiveVital}}. {In camera-based physiological sensing, physiological priors support self-supervised and weakly supervised signal extraction~\cite{Yue2023_SSL,Sun2024_ContrastPhys} and generalizable multi-task estimation~\cite{Wang2025_PhysMLE}.} Millimeter-wave (mmWave) radar offers a privacy-preserving modality for contactless monitoring that can sense chest-wall micro-motion through clothing with sub-millimeter displacement sensitivity~\cite{Hu2024}. {Its use in robotic interaction perception~\cite{Hu2026_WaveMan}, privacy-preserving human--robot interaction~\cite{Fan2026_mmHRI}, and fall detection~\cite{Lu2026_EMFall} illustrates its broader role in human sensing.} Extending radar-based physiological monitoring to everyday healthcare requires reliable heart-rate estimation~\cite{Rong2025}.}

\begin{figure}[t]
    \centering
    \includegraphics[width=\columnwidth]{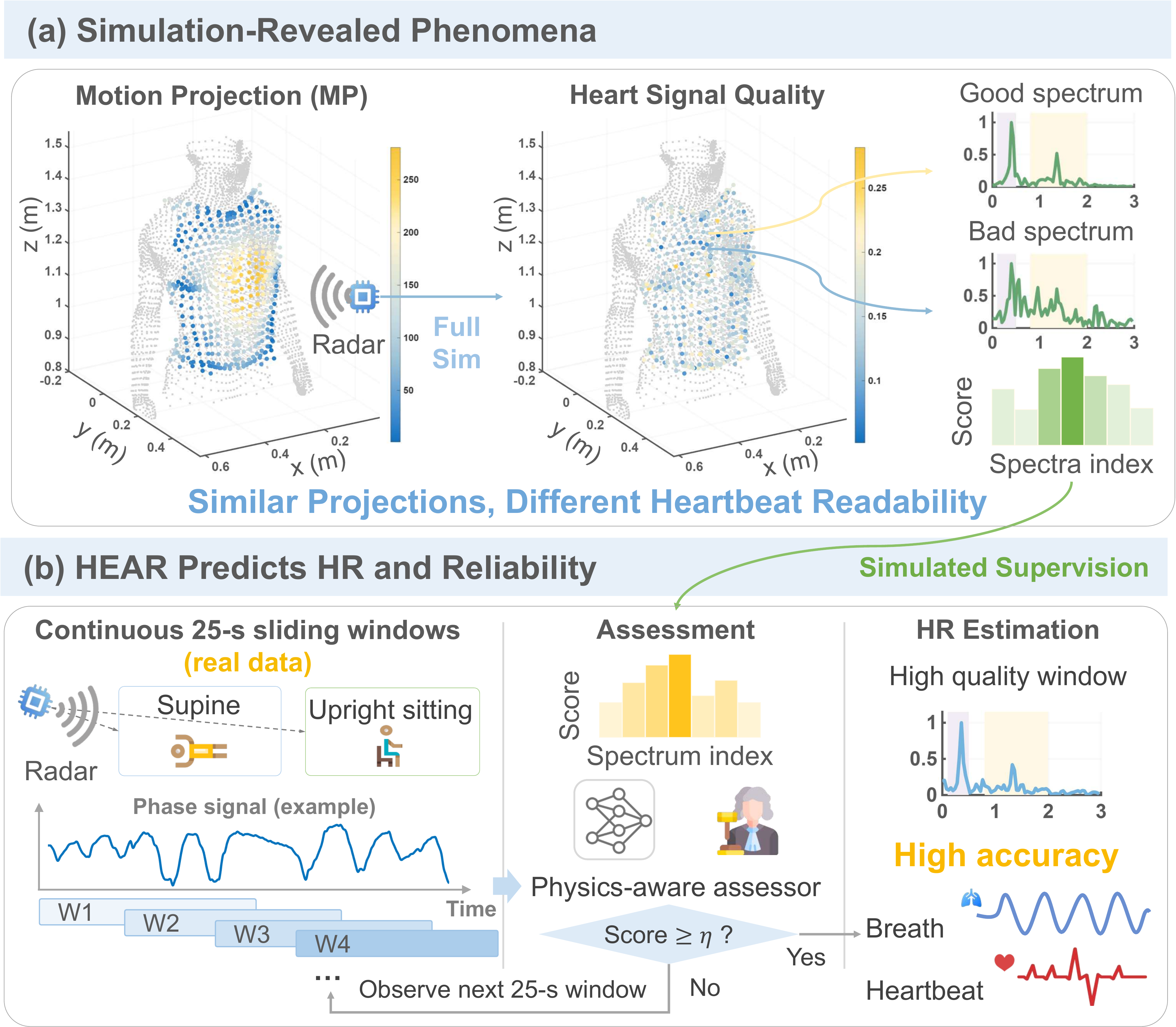}
    \caption{Motivation and overview of HEAR. (a) Simulated observations with similar motion projections can exhibit markedly different heartbeat observability, showing that geometry alone is insufficient to determine whether the heartbeat component is readable. This motivates assessment from the acquired signal. (b) Trained solely on simulated observations, HEAR is applied to real recordings without adaptation and jointly predicts an observability score and heart rate from the phase spectrum of each 25~s window. Heart-rate estimates are reported only for windows with predicted scores at or above the selection threshold.}
    \label{fig:teaser}
\end{figure}

{Despite recent advances, mmWave heart-rate estimation remains sensitive to environments and postures~\cite{Lu2026,Wu2026,Zhang2024}. Existing methods improve signal extraction through harmonic cancellation, mode decomposition, or learned regression~\cite{Lu2026,Yang2025,Seifizarei2025,Zhang2025}, or improve acquisition conditions through beamforming, multi-view placement, and orientation compensation~\cite{Antolinos2025,Ren2021_distributed,Gong2025_OSense,Zhang2026}. Yet an individual measurement may still lack a sufficiently distinct heartbeat component for reliable estimation.} We refer to this measurement-specific readability as \emph{heartbeat observability}. Many existing systems do not explicitly assess this property, making it difficult to distinguish reliable estimates from erroneous ones. {We study how to learn heartbeat observability from acquired signals and use the resulting score to select reliable measurements for different heart-rate estimators.}

{The need for measurement-level assessment has a physical basis.} Echoes from multiple body-surface scatterers superpose coherently within the same range bin~\cite{Lu2026,Zhang2024}. {Sub-wavelength differences in scatterer ranges alter their phase differences and can preserve or suppress the heartbeat component even under similar motion projections (Sec.~\ref{sec:theory}). Controlled simulations show markedly different heartbeat spectra at adjacent observation locations (Fig.~\ref{fig:teaser}a), consistent with location sensitivity reported in real measurements~\cite{Wu2026,Ren2021_distributed}. Perturbing only scatterer phases also changes observability while macroscopic geometry remains fixed (Sec.~\ref{sec:decouple}). Small body motions can introduce such phase changes over time. Favorable geometry therefore provides a useful prior, but the readability realized in each measurement must be assessed from the acquired signal.}

{Learning this assessment requires supervision across conditions in which the heartbeat component is preserved or obscured. Controlled acquisition of such observations is difficult because physiological motion varies across real recordings. We develop a multi-scatterer frequency-modulated continuous-wave (FMCW) simulator that generates paired observations under identical respiratory and heartbeat motion while varying observation geometry and scattering conditions. For each simulated spectrum, agreement between its dominant heartbeat-band peak and the known heart rate provides an automatic label of heartbeat readability. This supports learning a task-specific quality score from controlled measurements. A central challenge is whether a score learned from these simulated observations remains useful on real recordings acquired under different conditions. Prior work on cross-domain Wi-Fi gesture recognition demonstrates the value of physically motivated representations~\cite{Zhang2022_Widar3}. This motivates incorporating physiological frequency relationships into our spectral representation.}

{We address this challenge with HEAR (Heartbeat Estimation with Assessed Reliability), a compact dual-task Transformer trained entirely on simulated observations. Its input combines spectral magnitudes with frequencies expressed relative to the respiration fundamental, providing context for respiratory harmonics that may resemble heartbeat peaks. A single forward pass produces an observability score and a heart-rate estimate, and the score supports selective prediction with either HEAR's own head or an external estimator. We evaluate transfer without real-data adaptation on two public datasets collected at 60 and 120~GHz from 134 subjects. At 50\% coverage, selection reduces the HR head's MAE on the 120~GHz dataset from 17.9 to 1.6~BPM. The same score also improves selective estimation with multiple existing heart-rate estimators. The complete pipeline achieves an end-to-end processing latency of 50.8~ms on an edge device.}

{The main contributions are as follows.}

\begin{enumerate}[label=\arabic*)]
    \item {Physical basis for observability assessment. We show that coherent superposition of scatterer returns can change the heartbeat phase response under fixed motion projections. This finding establishes why macroscopic geometry alone is insufficient to determine heartbeat readability and motivates assessment from each acquired measurement.}

    \item {Controlled simulation for observability learning. We develop a multi-scatterer FMCW simulator that generates paired observations under identical respiratory and heartbeat motion across radar positions. The framework supports controlled analysis of phase-superposition effects and automatically derives spectral-readability labels, providing supervision across diverse observation conditions.}

    \item {Transferable observability assessment. We propose HEAR, a compact dual-task network that jointly predicts heartbeat observability and heart rate using spectral features with respiratory-harmonic context. Trained solely on simulated observations, HEAR transfers zero-shot to two real datasets. The same learned score supports selective prediction with both HEAR and multiple existing estimators, enabling an accuracy--coverage trade-off.}
\end{enumerate}

The rest of the paper is organized as follows. Sec.~\ref{sec:related} reviews related work, Sec.~\ref{sec:method} presents the proposed method, Sec.~\ref{sec:setup} describes the experimental protocol, Sec.~\ref{sec:exp} presents the results, Sec.~\ref{sec:discussion} discusses limitations, and Sec.~\ref{sec:expand} concludes the paper. The appendices provide additional implementation details, dataset descriptions, the segmentation protocol for zero-shot evaluation, and an analysis of segment-hop sensitivity.

\section{Related Work}\label{sec:related}

\begin{figure*}[t]
\centering
\includegraphics[width=\textwidth]{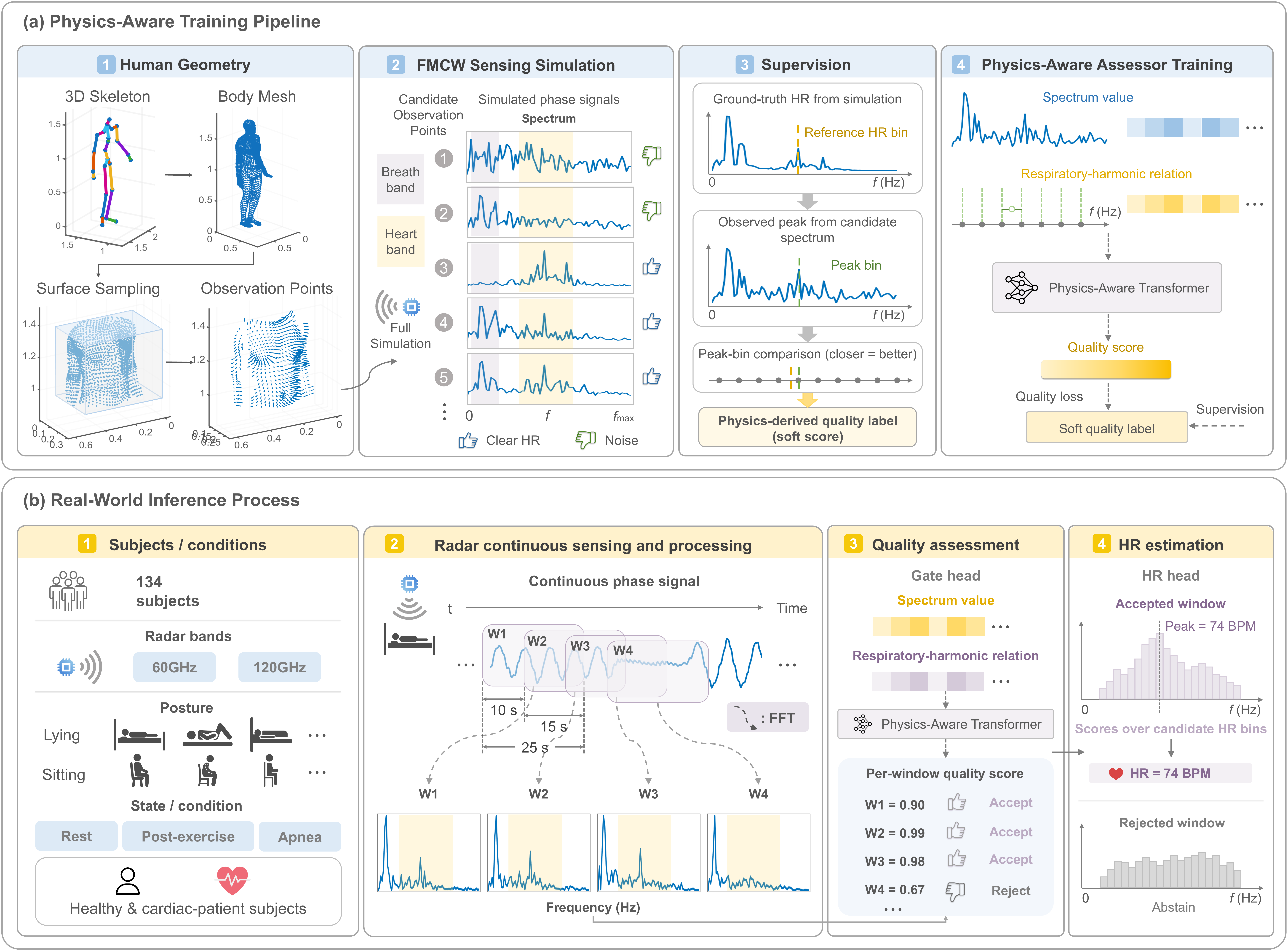}
\caption{Offline training and zero-shot inference with HEAR.
(a) The multi-scatterer FMCW simulator generates paired observations across radar positions under identical respiratory and heartbeat motion. The discrepancy between the dominant heartbeat-band peak and the reference heart rate defines a soft observability label. These labels and reference heart rates supervise joint observability prediction and HR estimation.
(b) Without real-data adaptation, HEAR processes the phase spectrum of each 25~s window, extracted with a 10~s hop, and jointly predicts an observability score and heart rate. The HR estimate is reported when the score meets or exceeds the selection threshold.}
\label{fig:overview}
\end{figure*}

\subsection{Radar-Based Heart-Rate Estimation}

Signal-processing approaches estimate heart rate from radar phase signals~\cite{Hu2024,Lu2026} through adaptive harmonic cancellation~\cite{Lu2026}, variational mode decomposition (VMD)~\cite{Yang2025}, autocorrelation~\cite{Seifizarei2025}, dictionary separation~\cite{Eder2026}, and multi-limb modeling~\cite{Ni2024}. Learning-based methods also reconstruct heartbeat waveforms~\cite{Wang2023} and electrocardiograms (ECGs)~\cite{Zhang2025,Chen2022_ECG,Lu2026_12leadECG} from radar measurements. Related work extends radar cardiac sensing to vital-motion imaging~\cite{Rong2025}, long-term cardiac monitoring~\cite{BB_Zhang2024_LongTerm}, arrhythmia detection~\cite{Yuan2025_AF,Cai2026_Arrhythmia}, and heart-rate variability (HRV) analysis~\cite{Wang2026_HRV}.

Approaches that account for observation geometry improve radar sensing through thorax-focused phased-array/multiple-input multiple-output (MIMO) beamforming~\cite{Antolinos2025,Li2024_4D,Ren2025_MultiVital}, distributed multi-angle acquisition~\cite{Ren2021_distributed}, mechanized translation/tilting~\cite{Zhang2025b}, and geometric compensation for non-normal incidence~\cite{Gong2025_OSense,Gong2024_Orientation}. Other methods select favorable observation points under signal feedback~\cite{Zhang2026} or adjust the radar pose under visual feedback~\cite{Hu2025_ActiveVital}. Studies also report substantial variation in heartbeat observability across measurement locations~\cite{Wu2026,Zhang2024,Ren2021_distributed}.

These approaches improve signal extraction or account for observation geometry, but estimation reliability can still vary across measurements. This motivates learning to assess whether an acquired signal supports reliable heart-rate estimation. HEAR learns this assessment from the phase spectrum and uses the resulting score to select measurements for both its own heart-rate head and existing estimators.

\subsection{Radar Signal Simulation and Human-Body Modeling}
Radar vital-sign simulation provides a controllable environment for studying radar signal formation. Early single-point models represent respiration and heartbeat as time-varying range~\cite{Xiong2017} but cannot capture multi-scatterer coherent superposition. Distributed models based on the Geometrical Theory of Diffraction (GTD) and physical optics discretize the body surface into scattering centers~\cite{Hu2024,Lu2026}, enabling analysis of radar returns and signal-to-noise characteristics. Motion-capture datasets such as AMASS~\cite{AMASS2019} and CMU MoCap~\cite{CMU_MoCap} provide resources for modeling human poses and movements. A recent SMPL-based shooting-and-bouncing-rays FMCW simulator~\cite{Li2024_SBRF,Hu2025_HAR} demonstrates sim-to-real transfer for millimeter-wave action recognition, but primarily models centimeter-scale body motion rather than sub-millimeter cardiac displacement.

These simulators primarily support signal modeling or motion recognition, leaving the effects of scatterer phases on heartbeat readability less explored. Controlled comparisons in real measurements are difficult because identical respiratory and heartbeat motion cannot readily be reproduced across observation locations~\cite{Xu2025_MassiveHRV}. We therefore develop a multi-scatterer simulator based on SMPL~\cite{Loper2015} that jointly models observation geometry, chest-wall micro-motion, and coherent phase superposition. It generates paired observations under identical respiratory and heartbeat motion, enabling controlled analysis of location-dependent observability. Agreement between the dominant heartbeat-band spectral peak and the known heart rate also provides automatic labels for training HEAR to assess heartbeat readability.

\subsection{Signal Quality Assessment for Physiological Sensing}

Signal quality assessment (SQA) has been widely studied in contact-based physiological sensing. Electrocardiogram (ECG) quality is assessed using statistical features and classifiers~\cite{Clifford2012,Behar2013} or unsupervised deep learning~\cite{Seeuws2022}, with similar approaches developed for photoplethysmography (PPG)~\cite{Mohagheghian2022}. Beyond computing a quality score, its usefulness can be evaluated by how well it predicts downstream task performance. This perspective is established in biometrics, where sample quality measures are evaluated through error-versus-reject analysis~\cite{Grother2007,Teixeira2017} and comparisons across recognition algorithms~\cite{Grother2007}.

In mmWave heart-rate estimation, several methods screen, select, or combine measurements using SNR thresholds, peak prominence~\cite{Lu2021}, periodicity~\cite{Seifizarei2025}, spectral energy ratios~\cite{Li2024_4D}, or SNR-based merging~\cite{Zhu2026}. These approaches primarily use predefined signal characteristics to guide measurement selection or combination. HEAR learns to assess heartbeat readability from phase-spectrum features, using simulation-derived labels based on agreement between the dominant heartbeat-band peak and the known heart rate. Following the task-performance perspective of quality assessment, we evaluate the score by whether it selects measurements that support accurate heart-rate estimation across different estimators.

\subsection{Selective Prediction}

Selective prediction augments a predictor with a selection rule that determines whether to retain its output. Its performance is evaluated through risk--coverage curves, which relate prediction error on retained samples to the fraction of samples retained~\cite{Chow1970,ElYaniv2010,Geifman2017,Geifman2019}. Selection commonly relies on reliability scores derived from maximum softmax probability~\cite{Hendrycks2017}, an auxiliary confidence-prediction network~\cite{Corbiere2019}, or predictive uncertainty estimated by deep ensembles~\cite{Lakshminarayanan2017}. These approaches motivate examining which scoring criteria effectively identify reliable predictions for a given task.

HEAR applies selective prediction to mmWave heart-rate estimation using a learned score of heartbeat spectral readability. Its supervision is derived from agreement between the dominant heartbeat-band peak and the known heart rate in simulation, without using the outputs or prediction errors of the HR head. This allows the same score to guide measurement selection for different estimators without redefining its training target for each one. We evaluate this use with HEAR's own HR head and existing classical estimators.

\begin{figure*}[!t]
\centering
\includegraphics[width=\textwidth]{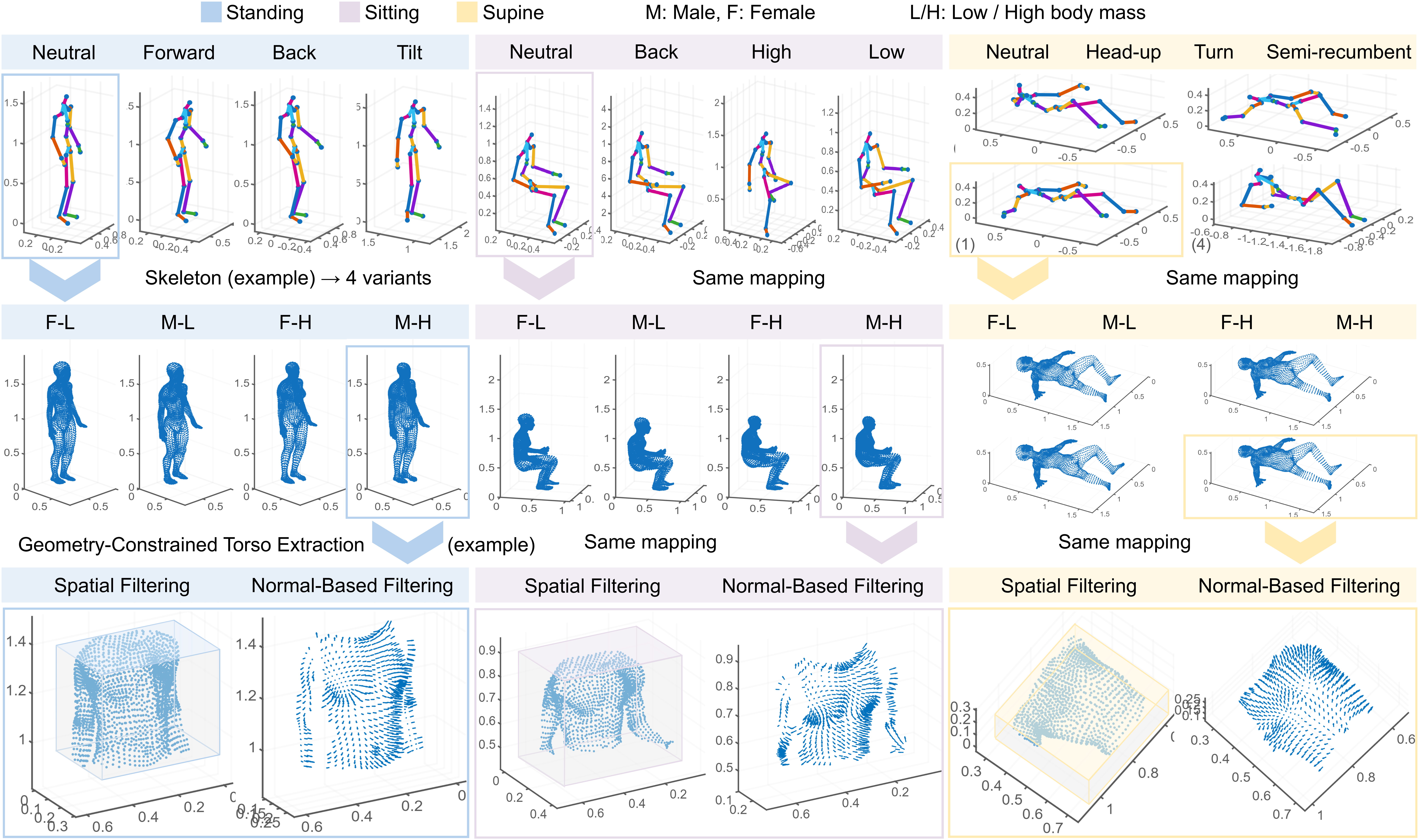}
\caption{SMPL body models and candidate radar position generation. Twelve standing, sitting, and lying postures are combined with four body types to construct 48 body models. Representative meshes illustrate variations in sex and body mass. Geometric constraints, spatial filtering, and surface-normal filtering identify candidate points on the anterior chest. Arrows indicate the surface normals used to place the radar at an outward offset of 0.40~m from each retained point.}
\label{fig:body}
\end{figure*}

\section{Method}\label{sec:method}
\subsection{Problem Definition and System Overview}\label{sec:problem}

Let $\phi_{\mathrm{d}}(m)$, $m=1,\dots,M$, denote the processed phase-difference signal of an observation segment, and let $X(b)$, $b=1,\dots,N_{\mathrm{b}}$, denote its magnitude spectrum at the $N_{\mathrm{b}}$ candidate heartbeat-frequency bins, used to construct the network input. HEAR jointly outputs an observability score $g\in[0,1]$ and a predicted HR distribution $\hat{\mathbf p}\in\mathbb{R}^{N_{\mathrm{b}}}$ over these bins, from which the heart-rate estimate $\hat f_{\mathrm{HR}}$ is obtained. The score estimates heartbeat readability in the acquired measurement, and the HR estimate is reported only when $g$ meets or exceeds a prescribed threshold.

Fig.~\ref{fig:overview} summarizes offline training and online inference. For training, we simulate radar observations across body geometries, radar positions, and respiratory and heartbeat motion, then derive observability labels and HR targets to supervise a physics-aware Transformer (Secs.~\ref{sec:sim} and~\ref{sec:net}). At inference, HEAR processes each acquired observation segment in a single forward pass and applies the selection rule described in Sec.~\ref{sec:infer}.
\subsection{Physical Basis for A Posteriori Observability Assessment}\label{sec:theory}

We analyze how coherent scattering affects the heartbeat-induced phase response and why macroscopic observation geometry alone cannot determine heartbeat observability. The key factor is the coherent superposition of echoes from multiple body-surface scatterers.

1) \emph{Multi-scatterer FMCW phase model.} An FMCW radar transmits a linear chirp. For a target at range $R$ with round-trip delay $\tau_{\mathrm{d}}=2R/c$, dechirping produces a beat signal containing a range-dependent frequency term and a propagation phase term $2\pi f_0\tau_{\mathrm{d}} = 4\pi R/\lambda$~\cite{Jankiraman}. Here, $c$ is the speed of light, $f_0$ is the chirp start frequency, and $\lambda=c/f_0$ is the corresponding wavelength. After range-bin selection, we model the phase response to chest-wall micro-motion using this dominant propagation term, such that a small range change $\Delta R$ produces a phase change $\Delta\phi\approx 4\pi\Delta R/\lambda$~\cite{Hu2025_ActiveVital,Hu2024}. This relation describes an individual scatterer. When multiple scatterers contribute to the same range bin, the measured phase is the argument of their summed complex returns, rather than the sum of their individual phases.

We select the maximum-energy chest bin for analysis. Following a scattering-center representation based on the Geometrical Theory of Diffraction (GTD)~\cite{Hu2024}, we model the return as a coherent sum of scatterer echoes. The selected range bin contains $N$ chest-wall scatterers at different distances from the radar, with sub-wavelength range differences producing different propagation phases. We denote the static range, nonnegative reflection magnitude, and propagation phase of scatterer $i$ by $R_i$, $a_i$, and $\psi_i=\beta R_i$, respectively, where $\beta = 4\pi/\lambda$ and $i=1,\dots,N$.

We model heartbeat motion using a common displacement waveform $s(t)$, with spatial variation represented by the weights $A_{\mathrm{heart}}(i)$. The effective weight $w_i=A_{\mathrm{heart}}(i)(\hat{\mathbf n}_i\cdot\hat{\mathbf u}_i)$ combines the local motion amplitude with its projection onto the radar line of sight, where $\hat{\mathbf n}_i$ is the outward unit surface normal at scatterer $i$ and $\hat{\mathbf u}_i$ is the unit line-of-sight vector from the radar toward the scatterer. Under a small-displacement approximation, the instantaneous range is $R_i(t)\approx R_i+w_i s(t)$, and the propagation phase is $\phi_i(t)=\psi_i+\beta w_i s(t)$. The slow-time complex signal at the target range bin is therefore
\begin{equation}\label{eq:super}
z(t)
=
\sum_{i=1}^{N} a_i\,e^{\,\mathrm{j}\phi_i(t)}
=
\sum_{i=1}^{N} a_i\,e^{\,\mathrm{j}\psi_i}\,
e^{\,\mathrm{j}\beta w_i s(t)} .
\end{equation}
In~\eqref{eq:super}, $e^{\,\mathrm{j}\psi_i}$ specifies each scatterer's baseline phase, while $e^{\,\mathrm{j}\beta w_i s(t)}$ describes its heartbeat-induced phase modulation. Equation~\eqref{eq:super} provides a simplified local phase model of the FMCW simulation in~\eqref{eq:ymix} and~\eqref{eq:ysum}. It retains the dominant propagation phase and treats scatterer amplitudes and their range-bin responses as fixed. The next subsection derives how these contributions combine to determine the heartbeat response of the measured phase.

2) \emph{Heartbeat modulation of the measured phase.}
To understand how coherent superposition can weaken the measured heartbeat response, we quantify how chest-wall displacement changes the phase of the combined return. Each scatterer contributes a displacement-dependent phase, but the measured phase is determined by their complex sum. We therefore examine the phase change per unit heartbeat displacement.

Writing~\eqref{eq:super} as a function of displacement $s$, let $\phi(s)$ denote the locally unwrapped phase of $z(s)$. Holding the scatterer amplitudes and projection weights fixed, direct differentiation gives
\begin{equation}\label{eq:z_local}
z'(s)
=
\mathrm{j}\beta
\sum_{i=1}^{N}
a_iw_i e^{\,\mathrm{j}\psi_i}
e^{\,\mathrm{j}\beta w_i s}.
\end{equation}
At the reference configuration $s=0$, define
\begin{equation}\label{eq:A0_G}
A_0
\triangleq
\sum_i a_i e^{\,\mathrm{j}\psi_i},
\qquad
G
\triangleq
\sum_i a_iw_i e^{\,\mathrm{j}\psi_i}.
\end{equation}
Here, $A_0=z(0)$ is the baseline complex return, while $G$ weights each scatterer return by its projected heartbeat motion, giving $z'(0)=\mathrm{j}\beta G$. These quantities describe the baseline return and its displacement response, respectively.

To extract the phase response from the complex-signal derivative, we use
\begin{equation}\label{eq:arg_identity}
\left.\frac{\mathrm{d}\phi}{\mathrm{d}s}\right|_{s=0}
=
\operatorname{Im}\!\left(\frac{z'(0)}{A_0}\right),
\qquad A_0\neq0.
\end{equation}
Substituting $z'(0)=\mathrm{j}\beta G$ yields
\begin{equation}\label{eq:phase_sensitivity}
\left.
\frac{\mathrm{d}\phi}{\mathrm{d}s}
\right|_{s=0}
=
\beta\,
\operatorname{Re}\!\left(\frac{G}{A_0}\right)
\triangleq
\beta h_\phi .
\end{equation}
The coefficient $h_\phi=\operatorname{Re}(G/A_0)$ characterizes the local sensitivity of the measured phase to heartbeat displacement. A smaller $|h_\phi|$ indicates a weaker phase response to an infinitesimal displacement around the reference configuration. When $h_\phi=0$, this local first-order response vanishes, even though individual scatterer phases may change. For adjacent samples near the reference configuration, where the phase sensitivity remains approximately constant, inter-frame differencing removes the baseline phase and gives $\Delta\phi\approx\beta h_\phi\Delta s$.

Equation~\eqref{eq:phase_sensitivity} follows directly from~\eqref{eq:super} without a small-phase expansion. It characterizes the response at the reference configuration. The sensitivity can vary over a finite heartbeat cycle. Because both $A_0$ and $G$ depend on the scatterer phases, we next examine how their relative phases affect this sensitivity under fixed motion projections.

3) \emph{Limits of geometry-based prediction.}
We examine whether macroscopic motion projection is sufficient to determine the local phase sensitivity $h_\phi$. For this analysis, we hold the reflection magnitudes $\{a_i\}$ and projection weights $\{w_i\}$ fixed within the signal model and vary the phase differences between scatterer echoes. The geometry-based projection score $Q_k=S\cdot C$ therefore remains unchanged, where $S=\sum_i|w_i|$ measures total projected motion and $C$ measures directional coherence. Any variation in $h_\phi$ under these conditions cannot be explained by $Q_k$.

The scatterer phases appear in both $G$ and $A_0$, so their effects may cancel in the ratio $G/A_0$. To determine when this cancellation occurs, we decompose each projection weight into an amplitude-weighted mean and a deviation from that mean. We define
\begin{equation}\label{eq:hetero_def}
\bar w
\triangleq
\frac{\sum_i a_iw_i}{\sum_i a_i},
\qquad
\delta w_i
\triangleq
w_i-\bar w,
\qquad
D
\triangleq
\sum_i a_i\delta w_i e^{\,\mathrm{j}\psi_i}.
\end{equation}
Substituting $w_i=\bar w+\delta w_i$ into $G$ gives
\begin{equation}\label{eq:hetero}
G=\bar w A_0+D,
\qquad
h_\phi
=
\bar w+\operatorname{Re}\!\left(\frac{D}{A_0}\right),
\quad A_0\neq0.
\end{equation}
The mean term $\bar w$ is independent of the scatterer phases. The correction $\operatorname{Re}(D/A_0)$ captures how differences in projected motion contribute to the phase response through coherent superposition. If all scatterers share the same projection weight, then $D=0$ and $h_\phi=\bar w$, independent of the phase differences. With heterogeneous projection weights, however, the correction can reinforce or offset the mean term as the phase differences change. Thus, identical motion projections and an unchanged $Q_k$ can yield different local phase sensitivities. The projection score alone is therefore insufficient to determine the local heartbeat response.

Only phase differences affect this result. A common phase shift multiplies $A_0$ and $D$ by the same factor and leaves their ratio unchanged. Physically, differences in propagation phase depend on differences in scatterer ranges through $\psi_i=4\pi R_i/\lambda$. Respiration and small body motions can alter these range differences even at a fixed radar pose, providing a mechanism for temporal variation in the phase response. Controlled simulations in Sec.~\ref{sec:decouple} and Fig.~\ref{fig:decouple} examine how scatterer phase variations affect heartbeat observability over finite observation windows.

4) \emph{Implications for a posteriori assessment.}
Macroscopic observation geometry does not specify the phase differences between scatterer echoes and therefore cannot uniquely determine the local heartbeat phase response. Heartbeat readability further depends on return amplitude, noise, respiratory interference, and signal processing. These factors jointly shape the acquired phase spectrum, motivating observability assessment directly from each measurement. The next subsection introduces a controlled simulation framework for investigating heartbeat observability across observation conditions.

\subsection{Simulation Data Generation}\label{sec:sim}

\begin{figure*}[!t]
\centering
\includegraphics[width=\textwidth]{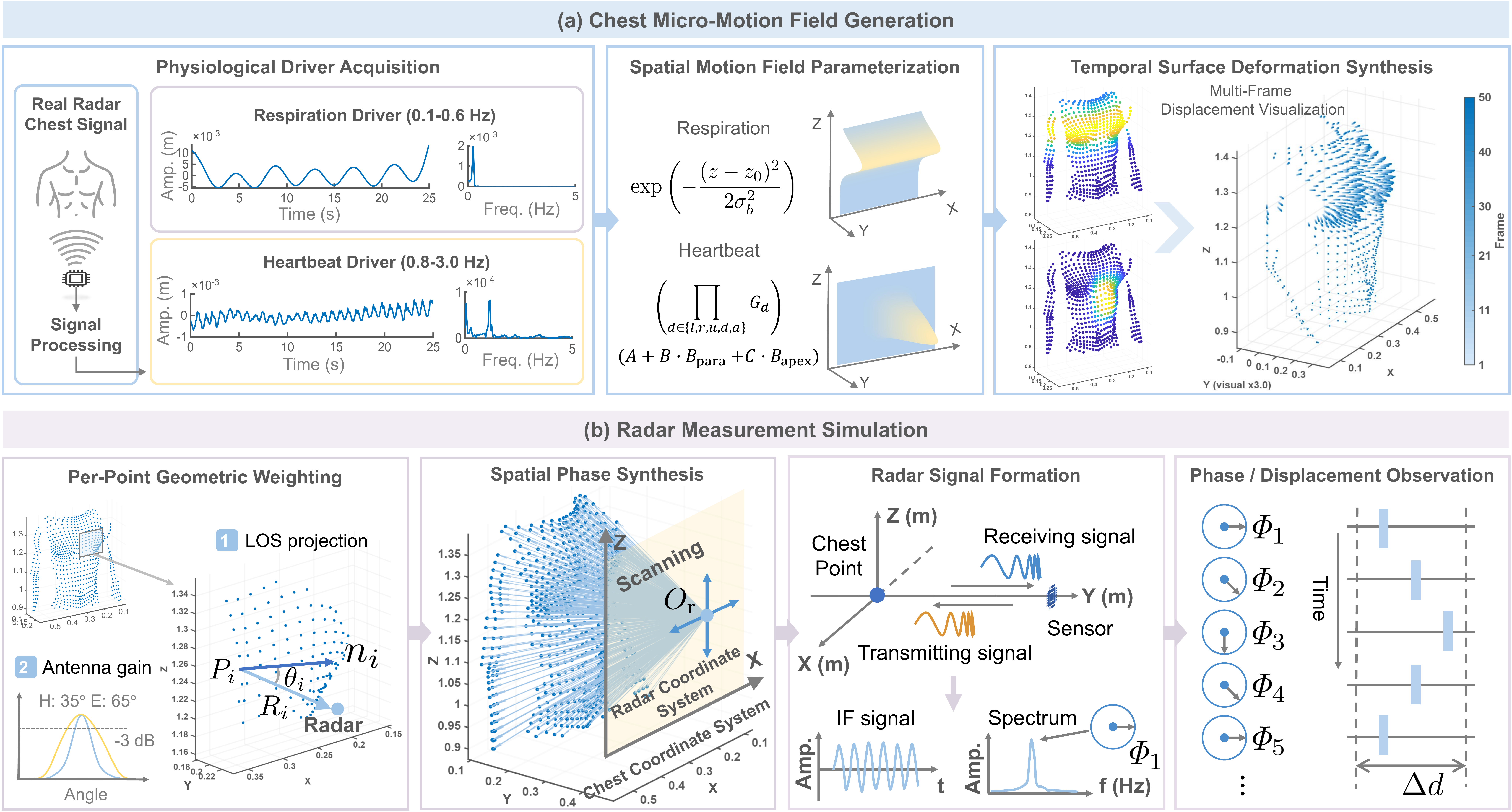}
\caption{Chest-wall motion synthesis and FMCW radar simulation.
(a) Chest-wall micro-motion synthesis combines respiration and heartbeat drive waveforms with spatial motion weights to generate displacement along each scatterer's surface normal over time.
(b) Radar measurement simulation uses scatterer ranges and antenna-pattern weights to generate per-scatterer FMCW mixed signals. These signals are coherently summed, combined with noise, and processed to extract the phase at the selected range bin. Simulated heartbeat frequencies span $[0.8,3.0]$~Hz, while the deployed HR output band is $[0.8,2.0)$~Hz.}
\label{fig:sim}
\end{figure*}

To study how observation conditions affect heartbeat observability, we need to compare radar measurements under the same respiratory and heartbeat motion and examine how individual scatterer echoes combine. Such comparisons are difficult in physical experiments because physiological motion varies across recordings and individual scatterer contributions are not directly available. We therefore develop a simulator that generates measurements at different radar positions and orientations under identical body motion and provides access to each scatterer's contribution. The known heart rates and simulated spectra also allow us to derive observability labels for training HEAR.

We first construct a library of 2{,}240 paired respiratory and heartbeat waveforms and 48 SMPL body models spanning 12 postures and 4 body types. We then apply the waveforms to the body surfaces using spatial motion weights to synthesize chest-wall displacement. Next, we apply geometric, spatial, and surface-normal constraints to identify approximately 600 eligible points on the anterior chest of each body model. We use these points to define candidate radar positions and retain the top 10\% based on projection quality. For each combination of body model and motion waveforms, we simulate the FMCW signal chain across the retained configurations. This produces paired measurements with identical underlying motion but different observation geometries. Finally, phase processing yields the signals used to construct network inputs, while the corresponding heart-rate references support label construction.

\subsubsection{Drive Signal Library}

We construct a library of respiratory and heartbeat waveforms from self-collected mmWave radar recordings to generate diverse chest-wall motion. We first divide the recordings into non-overlapping 25~s segments and retain stationary measurements with clear heartbeat content. The acquisition parameters and screening criteria are provided in Appendix~\ref{app:impl}. Respiratory and heartbeat signals extracted from the retained segments serve as motion drives for the simulator.

The extracted heartbeat drives are concentrated near resting heart rates. To broaden their frequency coverage, we shift their dominant frequencies to target values uniformly spaced on a 0.04~Hz grid over $[0.8,3.0]$~Hz. For a source heartbeat drive $d_{\mathrm{heart}}(t)$, we construct its analytic signal using the Hilbert transform and apply a frequency shift~{\cite{Weaver1956}} to obtain
\begin{equation}
d_{\mathrm{shifted}}(t)
=
\operatorname{Re}\!\left\{
\left[
d_{\mathrm{heart}}(t)
+
\mathrm{j}\mathcal{H}\{d_{\mathrm{heart}}(t)\}
\right]
e^{\,\mathrm{j}2\pi\Delta f t}
\right\}.
\end{equation}
Here, $\mathcal{H}\{\cdot\}$ denotes the Hilbert transform, and $\Delta f=f_{\mathrm{target}}-f_{\mathrm{source}}$ is the difference between the target frequency and the source drive's dominant frequency. Multiplication by the complex exponential shifts the analytic signal in frequency while preserving its amplitude envelope.

This augmentation yields 2{,}240 heartbeat drives. We then augment the respiration drives over the target breathing band and pair each heartbeat drive with a respiration drive from a different source recording. {During augmentation, drive amplitudes are additionally jittered by uniform random factors to increase physiological variability (Appendix~\ref{app:impl}).} These paired drives are subsequently combined with body models and radar positions to generate simulated observations.

\subsubsection{SMPL Scattering Model and Motion Synthesis}

We first construct the body-surface scattering geometry using the SMPL parametric human mesh~\cite{Loper2015,Fan2026_M4Human} (Fig.~\ref{fig:body}). SMPL-based meshes also provide supervision for Wi-Fi-based human reconstruction~\cite{11556388}. The mesh is controlled by pose parameters $\boldsymbol{\theta}\in\mathbb{R}^{72}$ and shape parameters $\boldsymbol{\beta}\in\mathbb{R}^{10}$. Each vertex is treated as a scatterer with a reference position $\mathbf{p}_i^0\in\mathbb{R}^3$ and an outward unit normal $\hat{\mathbf n}_i\in\mathbb{R}^3$. Combining 12 postures, including 4 standing, 4 sitting, and 4 lying postures selected from the CMU MoCap subset of AMASS~\cite{AMASS2019,CMU_MoCap}, with 4 body types yields 48 body models.

We then define spatial weights to distribute respiratory and heartbeat motion over the thoracic surface (Fig.~\ref{fig:sim}). The respiration weight follows a one-dimensional Gaussian centered on the upper chest,
\begin{equation}\label{eq:abreath}
A_{\mathrm{breath}}(i)
=
\exp\!\left[
-\frac{(\zeta_i-\zeta_c)^2}{2\sigma_b^2}
\right],
\end{equation}
where $\zeta_i$ is the coordinate of vertex $i$ along the torso-height direction, $\zeta_c$ is located at 75\% of the torso height, and $\sigma_b=0.15\,\Delta\zeta$, with $\Delta\zeta$ denoting torso height.

For heartbeat motion, we use directional sigmoid gates $G_q(i)$ to define the precordial region and Gaussian functions $B_{\mathrm{para}}(i)$ and $B_{\mathrm{apex}}(i)$ to increase the weights near the parasternal and apical regions~{\cite{Munck2020}}. The resulting weight is
\begin{equation}\label{eq:aheart}
\begin{aligned}
A_{\mathrm{heart}}(i) &= \prod_{q\in\{\ell,r,u,d,a\}}G_q(i)\\
&\quad\cdot\left[1+0.3\,B_{\mathrm{para}}(i)+0.3\,B_{\mathrm{apex}}(i)\right],
\end{aligned}
\end{equation}
where $q\in\{\ell,r,u,d,a\}$ denotes the left, right, upper, lower, and anterior constraints. This spatial pattern reflects the reported distribution of cardiac vibrations over the precordium~\cite{Munck2020}. The two Gaussian terms provide local enhancements with coefficients of 0.3. We subsequently normalize each spatial weight function to unit mean. The normalized weights determine the relative distribution of motion, while the displacement waveforms set its overall amplitude.

\begin{figure}[!t]
\centering
\includegraphics[width=\columnwidth]{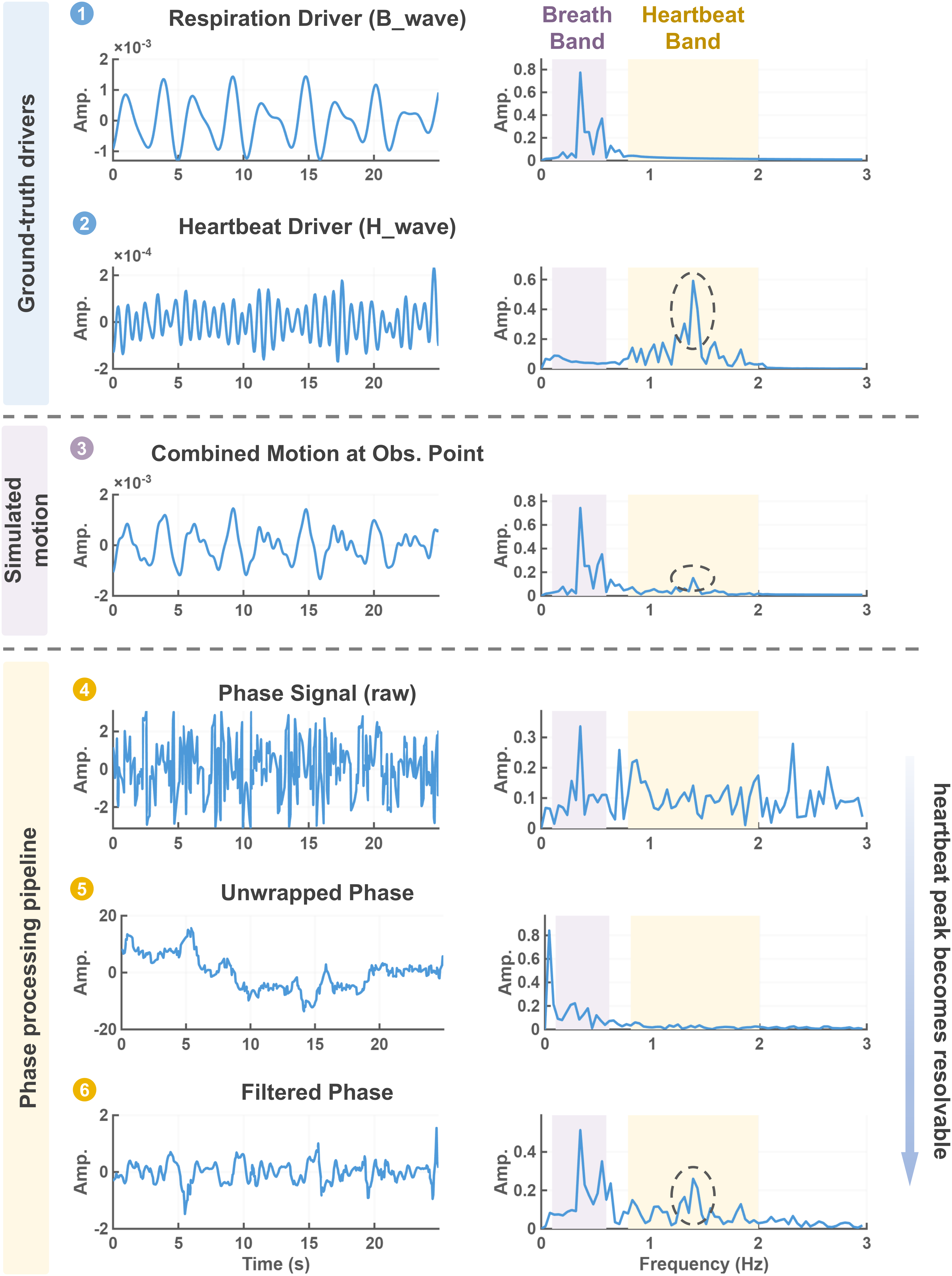}
\caption{Signal-processing chain for a simulated observation at one radar position.
Time-domain signals are shown on the left and spectra on the right, with respiration and heartbeat bands marked. The synthesized surface motion is used in FMCW simulation to generate the complex return at the selected range bin. Its phase is extracted, unwrapped, differenced between frames, Hampel-filtered, and smoothed. In this example, preprocessing makes the heartbeat peak more distinct in the spectrum of the processed phase-difference signal $\phi_{\mathrm{d}}$. This spectrum is used to construct the network input.}
\label{fig:chain}
\end{figure}

\begin{figure*}[!t]
\centering
\includegraphics[width=\textwidth]{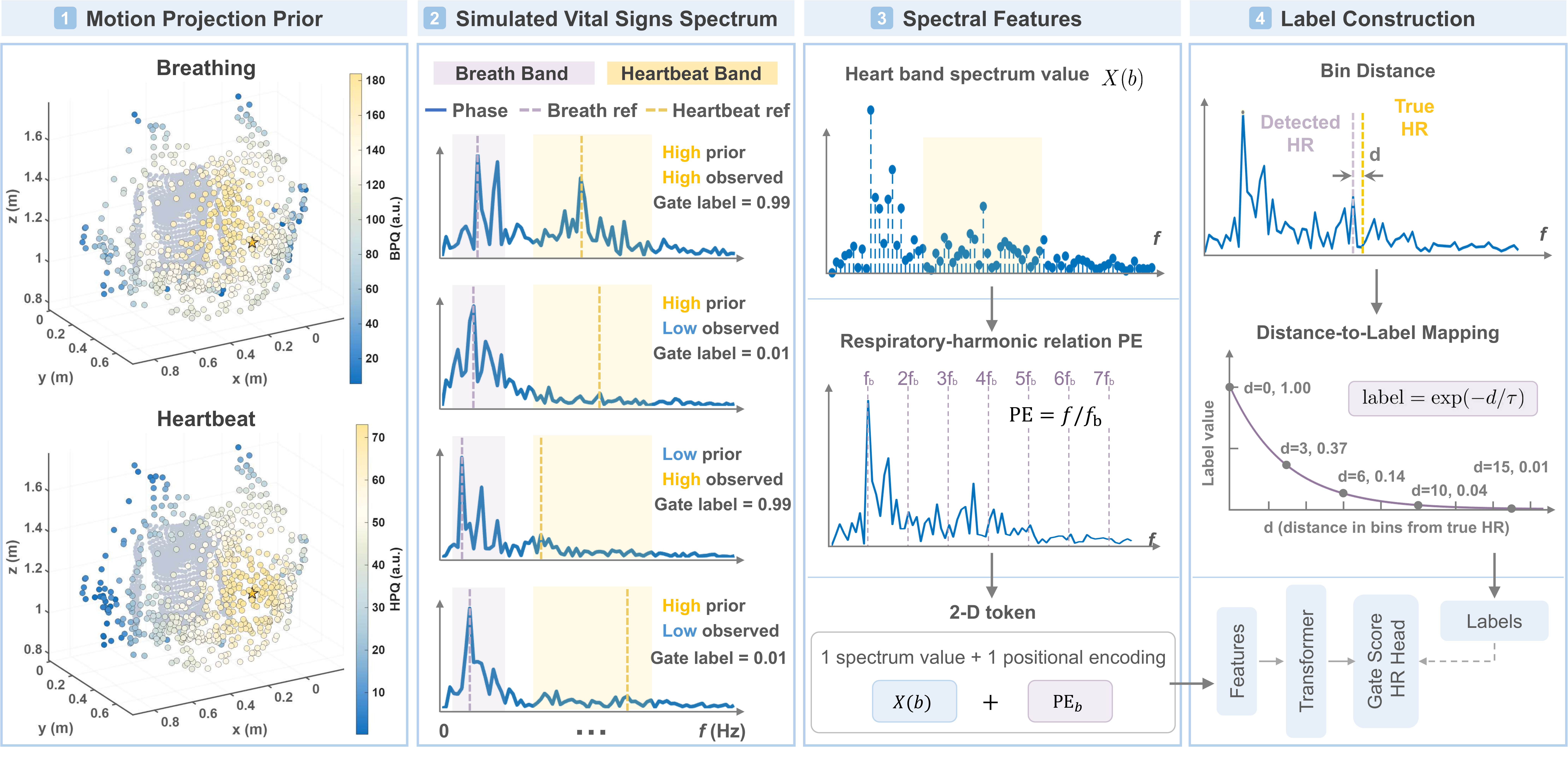}
\caption{Network inputs and observability-label construction.
Motion-projection maps and simulated phase spectra illustrate differences between geometric favorability and heartbeat observability. The projection scores are not network inputs. For each candidate heartbeat bin, HEAR forms a 2-D token $[X(b),r(b)]$, combining the normalized spectral magnitude with the frequency ratio $r(b)=f(b)/f_{\mathrm{b}}$, where $f_{\mathrm{b}}$ is the estimated respiration fundamental. The soft observability label is derived from the bin distance $d=|b_{\mathrm{peak}}-b_{\mathrm{true}}|$ between the dominant heartbeat-band peak and the reference HR, using $y_{\mathrm{gate}}=\exp(-d/\tau)$ with $\tau=3$.}
\label{fig:feat}
\end{figure*}

Next, we cumulatively sum each phase-difference drive to obtain a waveform for motion synthesis,
\begin{equation}
\widetilde B(m) = \sum_{k=1}^{m} d_{\mathrm{breath}}(k),
\end{equation}
where $m$ indexes frames. We obtain $\widetilde H(m)$ analogously from $d_{\mathrm{heart}}(k)$. {These waveforms are then converted from phase to displacement by the $\lambda/(4\pi)$ factor of the source radar, yielding $B_{\mathrm{wave}}(m)$ and $H_{\mathrm{wave}}(m)$. This conversion preserves the recorded amplitude scale.}

Finally, we combine the displacement waveforms with the normalized spatial weights to update each scatterer's position,
\begin{equation}\label{eq:motion}
\mathbf{p}_i(m) = \mathbf{p}_i^0 + \hat{\mathbf n}_i
\left[
A_{\mathrm{breath}}(i) B_{\mathrm{wave}}(m)
+
A_{\mathrm{heart}}(i) H_{\mathrm{wave}}(m)
\right].
\end{equation}
Each scatterer thus moves along its reference surface normal, with respiratory and heartbeat displacement determined by the corresponding waveform and spatial weight.

\subsubsection{Candidate Radar Position Generation} \label{sec:III-C-4}

For each body model, we first extract the thoracic region using geometric constraints. We then apply spatial and surface-normal filters to retain points on the anterior chest with forward-facing surface normals. This yields approximately 600 candidate surface points per body (Fig.~\ref{fig:body}). For each retained point $\mathbf{p}_k$ with outward unit normal $\hat{\mathbf n}_k$, we place the radar at an outward offset of $d_{\mathrm{offset}} = 0.40$~m and direct its boresight toward $\mathbf{p}_k$. The resulting radar position is
\begin{equation}
\mathbf{r}_k = \mathbf{p}_k + d_{\mathrm{offset}} \hat{\mathbf n}_k .
\end{equation}

We then evaluate the motion-projection geometry at each candidate position. For position $k$, let $\theta_i$ denote the angle between the outward normal $\hat{\mathbf n}_i$ and the direction from the radar to scatterer $i$, consistent with the convention in Sec.~\ref{sec:theory}. We define
\begin{equation}\label{eq:qk}
\begin{aligned}
c_i &= A_{\mathrm{heart}}(i)\cos\theta_i,
&
S &= \sum_i |c_i|,\\
C &= \frac{\left| \sum_i c_i \right|}{\sum_i |c_i|},
&
Q_k &= S \cdot C .
\end{aligned}
\end{equation}
Here, $c_i$ is the signed projection weight of heartbeat motion at scatterer $i$. The quantity $S$ measures the total magnitude of these weights, while $C$ measures their directional consistency. A larger $Q_k$ therefore favors positions with stronger and more consistent motion projections.

Finally, we rank the candidate positions by $Q_k$ and retain the top 10\% for simulation data generation. As shown in Sec.~\ref{sec:theory}, this score describes motion projection but does not account for the phase differences between scatterer echoes. It is used only to select candidate positions. Observability labels are subsequently derived from the simulated spectra and known heart rates, as described in Sec.~\ref{sec:net}.

\subsubsection{FMCW Radar Simulation and Phase Processing}

For each retained radar position $\mathbf{r}_k$, we simulate the FMCW signal chain using the parameters listed in Appendix~\ref{app:impl}. Each frame contains one chirp. For scatterer $i$ at frame $m$, the mixed signal is
\begin{equation}\label{eq:ymix}
\begin{aligned}
y_{\mathrm{mix}}^{(i)}(t,m)
=\gamma_i\exp\Big\{\mathrm{j}\Big[
&2\pi\big(
f_0\tau_{\mathrm{d},i}(m)
+K_{\mathrm{s}}\tau_{\mathrm{d},i}(m)\,t\\
&\quad+f_{\mathrm{d},i}(m)\,t
\big)
-\pi K_{\mathrm{s}}\tau_{\mathrm{d},i}^{2}(m)
\Big]\Big\},
\end{aligned}
\end{equation}
where $t$ denotes fast time within a chirp, $K_{\mathrm{s}}$ is the chirp slope, and $\tau_{\mathrm{d},i}(m)=2R_i(m)/c$ is the round-trip delay. The scatterer range is $R_i(m)=\|\mathbf{p}_i(m)-\mathbf{r}_k\|$, and the Doppler frequency is $f_{\mathrm{d},i}(m)=2v_{r,i}(m)/\lambda$, with $v_{r,i}(m)$ denoting radial velocity. The receive antenna-pattern weight $\gamma_i$ is modeled by a Gaussian beam and sets the relative scatterer amplitude under a uniform-material assumption.

We then coherently sum the scatterer returns and add complex Gaussian noise,
\begin{equation}\label{eq:ysum}
y(t, m) = \frac{1}{N} \sum_{i=1}^{N} y_{\mathrm{mix}}^{(i)}(t, m) + n(t,m),
\end{equation}
where $n(t,m) \sim \mathcal{CN}(0, \sigma_n^2)$. The noise level is specified in Appendix~\ref{app:impl}.

Next, we apply a range fast Fourier transform (FFT) along fast time and select the maximum-energy chest bin. We extract its complex sequence across frames, compute and unwrap its phase, and apply inter-frame differencing, Hampel filtering, and moving-average smoothing. This produces the processed phase-difference signal $\phi_{\mathrm{d}}(m)$, from which the spectral input is constructed. Although~\eqref{eq:ymix} includes the Doppler contribution within each chirp, the subsequent processing extracts heartbeat information from phase variations across frames without estimating Doppler frequency.

Repeating this procedure across body models, motion waveforms, and retained radar positions produces paired observations under controlled conditions. These observations support the analysis of phase-superposition effects and provide the phase signals and heart-rate references used to construct network inputs and training labels in Sec.~\ref{sec:net}.

\subsection{Physics-Aware HEAR Network}\label{sec:net}

HEAR uses a compact dual-task Transformer to jointly estimate heartbeat observability and heart rate from the acquired phase spectrum. Its input features combine spectral magnitude with frequency information relative to respiration, allowing the network to consider respiratory harmonics when interpreting candidate heartbeat peaks.

\begin{figure*}[!t]
\centering
\includegraphics[width=\textwidth]{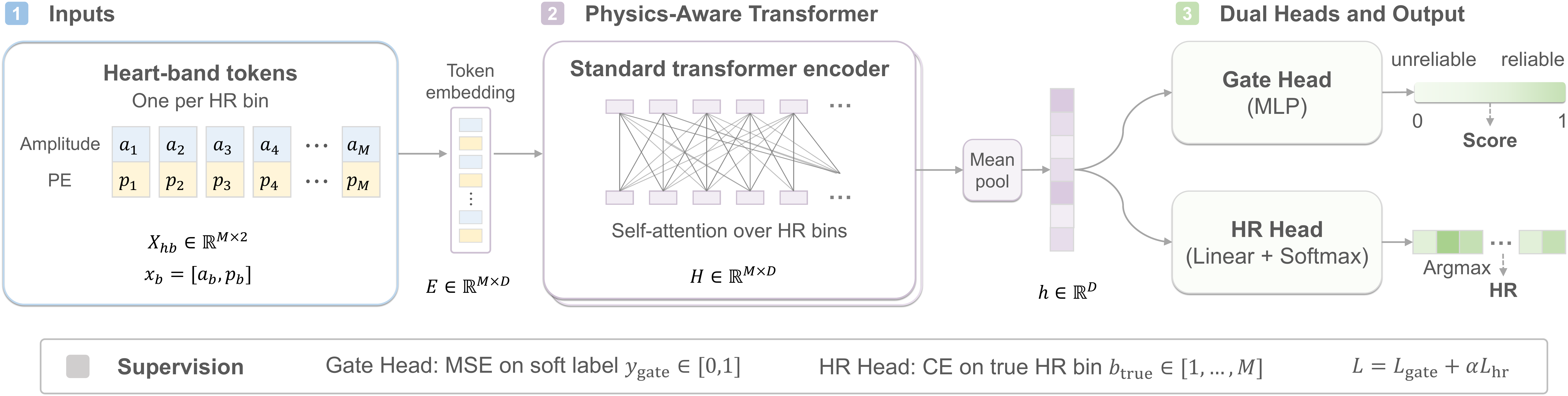}
\caption{Physics-aware dual-task network.
HEAR represents each candidate heartbeat bin by a 2-D token $[X(b),r(b)]$, containing the normalized spectral magnitude and the frequency ratio $r(b)=f(b)/f_{\mathrm{b}}$. A lightweight Transformer models relations across HR bins, and its output tokens are mean-pooled into a shared representation. Two heads jointly predict the observability score $g$ and the HR posterior $\hat{\mathbf p}$. The observability head is supervised by soft labels for all training observations, while the HR head is supervised by reference HR bins only for observations within the deployed HR band.}
\label{fig:net}
\end{figure*}

\subsubsection{Input Token Construction}

We first compute the magnitude spectrum of the processed phase-difference signal $\phi_{\mathrm{d}}(m)$, retain the 75 bins within $[0,3.0)$~Hz, and apply $L^2$ normalization over these bins. We estimate the respiration fundamental $f_{\mathrm{b}}$ as the frequency of the largest spectral peak within $[0.1,0.6]$~Hz on the 0.04~Hz grid. We then select candidate heartbeat frequencies within $[0.8,2.0)$~Hz, yielding $N_{\mathrm{b}}=30$ bins. For bin $b$, let $f(b)$ denote its frequency and $X(b)$ its normalized spectral magnitude.

Spectral magnitude alone does not reliably distinguish heartbeat peaks from respiratory harmonics. Non-sinusoidal respiratory motion and nonlinear phase extraction can produce harmonics at integer multiples of $f_{\mathrm{b}}$, some of which fall within the heartbeat band. To provide this context, we represent each candidate bin with a two-dimensional token,
\begin{equation}
\mathbf{x}_b = [X(b),\; r(b)], \qquad r(b)=\frac{f(b)}{f_{\mathrm{b}}}.
\end{equation}
The ratio $r(b)$ expresses the candidate frequency relative to the estimated respiration fundamental. Its relation to respiratory harmonics can be seen from the distance to the nearest harmonic,
\begin{equation}\label{eq:harm}
d_{\mathrm{harm}}(b)
= \min_{n\in\mathbb{Z}^+}\big|f(b)-n f_{\mathrm{b}}\big|
= f_{\mathrm{b}}\,\big|r(b)-\operatorname{round}(r(b))\big|.
\end{equation}
Thus, a ratio near an integer indicates that the candidate frequency lies close to a respiratory harmonic. Combined with the spectral magnitude $X(b)$, this information allows HEAR to assess candidate peaks in the context of both their strength and their relation to respiration. The network receives the tokens $[X(b),r(b)]$, while $d_{\mathrm{harm}}(b)$ describes the harmonic relationship encoded by the ratio.

\subsubsection{Observability-Label and HR-Target Construction}

We derive both training targets from the known heart rate and the simulated phase spectrum. We operationalize heartbeat observability through the agreement between the dominant spectral peak and the reference heart rate. This provides a measure of whether a peak-based estimator can correctly read the heartbeat frequency from the acquired spectrum.

For each observation, we first identify the dominant peak within the deployed heartbeat band,
\begin{equation}
b_{\mathrm{peak}}
=
\arg\max_{b\in\{1,\ldots,N_{\mathrm{b}}\}}X(b).
\end{equation}
Let $b_{\mathrm{true}}$ denote the reference heartbeat frequency expressed as an index on the same 0.04~Hz grid. For label construction, this grid extends beyond the deployed band to include all augmented heartbeat frequencies. We then define the continuous observability label as
\begin{equation}\label{eq:ygate}
y_{\mathrm{gate}}
=
\exp\!\left(
-\frac{|b_{\mathrm{peak}}-b_{\mathrm{true}}|}{\tau}
\right),
\end{equation}
where $\tau=3.0$ controls the decay with frequency-bin error. With a bin spacing of 0.04~Hz, a separation of three bins corresponds to 7.2~BPM and gives $y_{\mathrm{gate}}=e^{-1}$. The label equals 1 when the dominant peak matches the reference bin and decreases as their separation increases. This soft target retains the magnitude of the spectral mismatch rather than reducing it to a binary outcome.

For observations with reference frequencies in $[0.8,2.0)$~Hz, $b_{\mathrm{true}}$ also serves as the HR classification target. Augmented observations with reference frequencies in $[2.0,3.0]$~Hz fall outside the deployed band and supervise only the observability head. Their observability labels follow the same distance-based rule, so observations near the upper band boundary can retain intermediate or high labels.

\subsubsection{Transformer Backbone}

Each token $\mathbf{x}_b=[X(b), r(b)]$ is linearly embedded into a $d_{\mathrm{model}}=16$-dimensional feature and passed through a 2-layer Transformer backbone with 4 attention heads. No additional positional encoding is applied. Each bin's frequency coordinate is represented by the ratio $r(b)$ in its input token. Each layer uses standard scaled-dot-product multi-head attention~\cite{Vaswani2017}, followed by a Pre-LN feed-forward block with hidden dimension 32, GELU activation, and dropout 0.1. The output tokens are then mean-pooled to obtain a compact segment representation $\bar{\mathbf h}\in\mathbb{R}^{d_{\mathrm{model}}}$.

\subsubsection{Dual-Task Heads}

HEAR uses two heads on the pooled representation $\bar{\mathbf h}$. The gate head predicts the heartbeat observability score, and the HR head predicts the heartbeat-bin posterior,
\begin{align}
g &= \sigma\!\left( \mathbf W_2\,\mathrm{ReLU}(\mathbf W_1\bar{\mathbf h}) \right),\\
\hat{\mathbf p} &= \mathrm{softmax}\!\left( \mathbf W_{\mathrm{hr}}\bar{\mathbf h} \right),
\end{align}
where $\sigma$ is the sigmoid, $g\in[0,1]$ is the observability score, and $\hat{\mathbf p}\in\mathbb{R}^{N_{\mathrm{b}}}$ is the heart-rate posterior over the deployed heartbeat band. The learnable matrices are $\mathbf W_1\in\mathbb{R}^{d_{\mathrm{model}}\times d_{\mathrm{model}}}$, $\mathbf W_2\in\mathbb{R}^{1\times d_{\mathrm{model}}}$, and $\mathbf W_{\mathrm{hr}}\in\mathbb{R}^{N_{\mathrm{b}}\times d_{\mathrm{model}}}$. All linear layers include learnable biases, which are omitted from the equations for brevity. The network has 5{,}295 parameters in total.

\subsubsection{Loss and Training Configuration}

For each training observation, we define the HR loss as
\begin{equation}
\ell_{\mathrm{HR}}=
\begin{cases}
\mathrm{CE}(\hat{\mathbf p},b_{\mathrm{true}}),
& f_{\mathrm{true}}\in[0.8,2.0)~\mathrm{Hz},\\
0,&\text{otherwise},
\end{cases}
\end{equation}
where $f_{\mathrm{true}}$ is the reference heartbeat frequency. The joint loss for that observation is
\begin{equation}
\ell=(g-y_{\mathrm{gate}})^2+\alpha_{\mathrm{hr}}\ell_{\mathrm{HR}},
\end{equation}
with $\alpha_{\mathrm{hr}}=0.5$. All samples supervise the observability score, while only in-band samples contribute to HR classification training. The optimizer, learning-rate schedule, and other hyperparameters are given in Appendix~\ref{app:impl}.

\subsection{Observability-Guided Selective Inference}\label{sec:infer}

For each 25~s observation window, HEAR outputs the observability score $g$ and the HR distribution $\hat{\mathbf p}$ in a single forward pass. Given a selection threshold $\eta\in[0,1]$, the selective output is defined as
\begin{equation}
\hat{f}_{\mathrm{HR}} =
\begin{cases}
f\!\left(\arg\max_b \hat{\mathbf p}(b)\right), & g \geq \eta,\\
\mathrm{reject}, & g < \eta.
\end{cases}
\end{equation}
Accepted frequency estimates are converted from Hz to BPM by multiplying by 60. The same score can also select observation windows for external heart-rate estimators, such as FFT peak estimation (Sec.~\ref{sec:methodcmp}). Varying $\eta$ controls coverage and allows evaluation of the risk--coverage trade-off, as examined on real data in Sec.~\ref{sec:fusion}.

\begin{figure}[!t]
\centering
\includegraphics[width=\columnwidth]{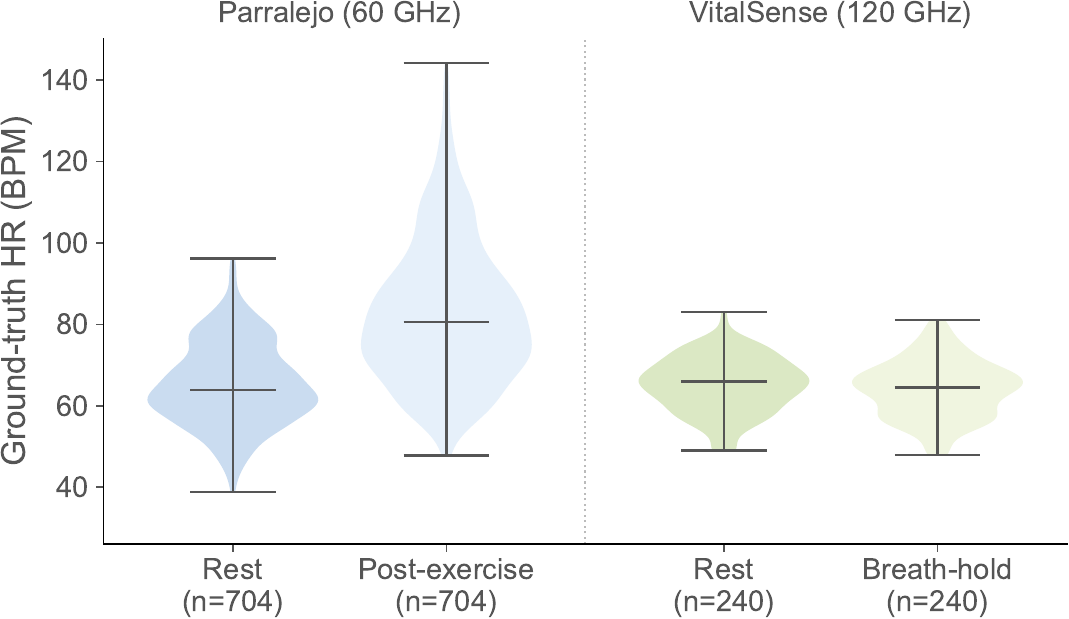}
\caption{Ground-truth HR distributions of sliding windows in the two public datasets, grouped by acquisition condition.
Each window spans 25~s with a 10~s hop, and $n$ denotes the number of windows in each group. In each violin, the central horizontal line marks the median, and the lower and upper caps indicate the minimum and maximum, respectively. Parralejo 60~GHz shows a broader HR distribution after exercise, reaching 144~BPM, whereas VitalSense 120~GHz remains entirely at or below 83~BPM under both resting and breath-hold conditions.}
\label{fig:hr_dist}
\end{figure}

\begin{figure*}[!t]
\centering
\subfloat[]{\includegraphics[width=0.24\textwidth]{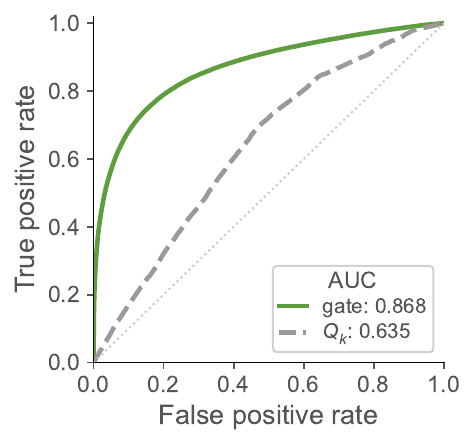}}\hfill
\subfloat[]{\includegraphics[width=0.24\textwidth]{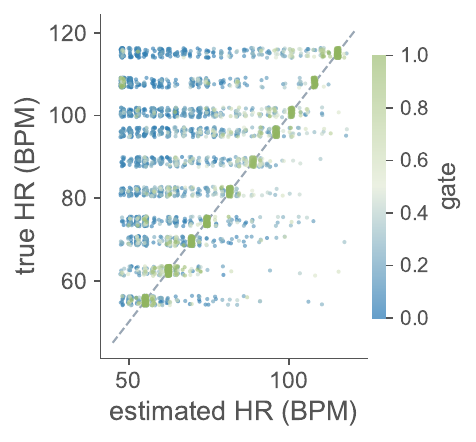}}\hfill
\subfloat[]{\includegraphics[width=0.25\textwidth]{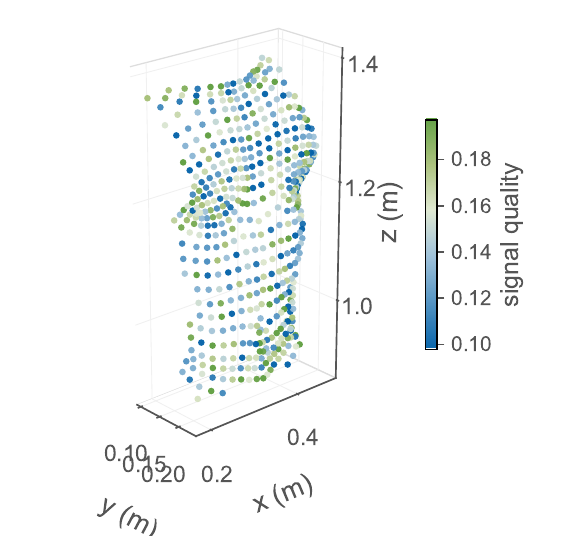}}\hfill
\subfloat[]{\includegraphics[width=0.24\textwidth]{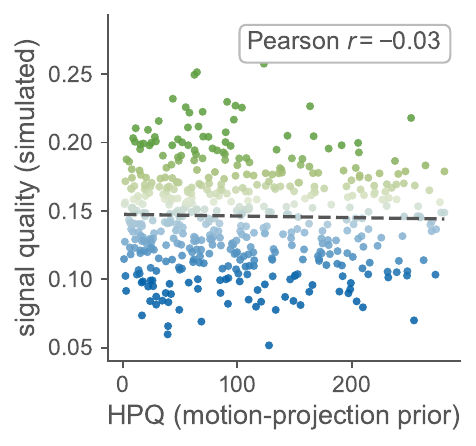}}
\caption{Geometry and heartbeat observability in simulation.
(a) ROC curves within the top-10\% projection-quality subset. HEAR and $Q_k$ achieve AUCs of 0.868 and 0.635, respectively.
(b) Dominant-peak HR estimates under differential sub-wavelength range perturbations, colored by observability score. The dashed line indicates perfect agreement. Splitting at the median score yields Acc@$\pm1$ bin of 85\% and 23\% for the higher- and lower-score groups.
(c) Heartbeat energy concentration across 492 positions on one body model. Color denotes the fraction of heartbeat-band energy within $\pm2$ bins of the reference HR.
(d) Heartbeat energy concentration versus $Q_k$ (Pearson $r=-0.03$).}
\label{fig:decouple}
\end{figure*}

\section{Experimental Protocol}\label{sec:setup}
\subsection{Datasets, Preprocessing, and Windowing}

The simulation set is generated using the procedure in Sec.~\ref{sec:sim}, with network inputs and labels constructed as described in Sec.~\ref{sec:net}. Combining 2{,}240 augmented heartbeat drives, each paired with a respiration drive, with 48 body models yields 107{,}520 body-drive pairs. Each pair contains about 50 retained observation positions, giving 5{,}631{,}052 observations in total. We assign 90\% of the pairs to training and 10\% to testing, keeping all positions from the same pair in one subset. The resulting test set contains 562{,}986 observations. This split evaluates internal consistency within the simulator because individual body models and augmented drives can occur in both subsets. Generalization to real measurements is evaluated separately.

We evaluate zero-shot transfer on two public FMCW datasets, Parralejo~\cite{Parralejo2026} at 60~GHz and VitalSense~\cite{Wu2026_VitalSense} at 120~GHz, containing 110 and 24 subjects, respectively. Both provide synchronized contact references for heart-rate evaluation. No window from either dataset is used for training, validation, model selection, or threshold tuning. All network training observations are simulated. The motion drives are extracted from separate self-collected recordings with no subject, device, or record overlap with the evaluation datasets, as detailed in Appendix~\ref{app:impl}.

All observations use a 25~s analysis window. Real recordings are segmented with a 10~s hop, yielding 1{,}408 Parralejo windows and 480 VitalSense windows. Simulated and real observations are converted into the normalized phase spectra used by HEAR. Dataset-specific preprocessing and reference heart-rate construction are described in Appendix~\ref{app:dataset}.

\subsection{Baselines, Metrics, and Selective Evaluation}

We first compare reliability scores. The baselines include hand-crafted spectral-shape and respiratory-harmonic indicators, together with two learned assessors, MLP-on-spectrum and 1D-CNN (Table~\ref{tab:bench}). The hand-crafted indicators measure heart-band peak strength, spectral concentration, peak prominence, spectral flatness, frequency ratio, harmonic distance, harmonic order, and harmonic proximity. The geometry-derived projection score $Q_k$ is evaluated only in simulation, where body geometry is available. We also test whether the same HEAR gate can select useful observations for different HR estimators. These include HEAR's own HR head, FFT peak, weighted FFT peak, autocorrelation~\cite{Seifizarei2025}, harmonic product spectrum, and VMD~\cite{Yang2025} (Table~\ref{tab:methodcmp}). The weighted FFT peak uses the energy centroid within $\pm3$ bins around the FFT maximum.

Gate-discrimination AUC evaluates whether a score distinguishes observations whose dominant heartbeat-band peak agrees with the reference heart rate. A window is positive when the peak lies within $\pm1$ bin of the reference, corresponding to $\pm0.04$~Hz or $\pm2.4$~BPM. This criterion is the binary counterpart of the soft label in Eq.~\eqref{eq:ygate}. We also report $\mathrm{Spearman}(g,-|err_{\mathrm{head}}|)$, where $err_{\mathrm{head}}$ is the error of HEAR's HR-head estimate, to assess whether higher scores correspond to lower estimation errors.

HR performance is measured by MAE and accuracy on the retained windows. Acc@5 BPM is the percentage of estimates within 5~BPM of the reference. For the cross-estimator comparison in Table~\ref{tab:methodcmp}, simulation accuracy uses a tolerance of $\pm1$ bin, while real-data accuracy uses $\pm5$~BPM. Coverage is the fraction of evaluation windows retained by the selection rule.

We use two selective evaluation settings. For fixed-coverage comparisons, each scoring method ranks windows and retains the highest-scoring fraction. All selectors use the same HEAR HR-head estimates, allowing comparison of selection quality at matched coverage. An oracle selector ranks windows by their reference-based absolute errors and provides a lower-risk benchmark. We summarize the risk--coverage curves using normalized AURC~\cite{Geifman2019}, computed by trapezoidal integration over 96 evenly spaced coverage levels from 5\% to 100\% and divided by the 0.95 coverage span. Lower AURC indicates better selective performance. For cross-estimator evaluation, we apply the fixed threshold $g\geq0.7$ and use the same retained windows for every estimator within each dataset.

\section{Experimental Results}\label{sec:exp}
\subsection{Geometry and Heartbeat Observability}\label{sec:decouple}

We first examine whether the motion-projection score $Q_k$ identifies observations with a reliable heartbeat peak. We treat observations whose dominant heartbeat-band peak lies within $\pm1$ bin of the reference heart rate as reliable and use this criterion to compute AUC. Among the positions retained by the top-10\% projection-quality screening, $Q_k$ achieves an AUC of 0.635, compared with 0.868 for the learned gate (Fig.~\ref{fig:decouple}(a)). This result indicates limited discrimination by $Q_k$ within the geometrically favorable subset. It does not characterize its performance over all candidate positions.

Fig.~\ref{fig:decouple}(b) examines phase sensitivity through differential sub-wavelength line-of-sight perturbations. For each reference heartbeat frequency, we hold the macroscopic radar position fixed while applying the perturbations. The perturbations alter the relative static phases $\{\psi_i-\psi_j\}$ while leaving $Q_k$ effectively unchanged. The dominant-peak HR estimates vary widely across the perturbed observations, showing differences in spectral readability despite similar motion projections. This behavior is consistent with the coherent-superposition mechanism analyzed in Sec.~\ref{sec:theory}.

Panels (c)--(d) examine heartbeat energy concentration across 492 evaluated candidate positions on one body model. We define this quantity as the fraction of total heartbeat-band spectral energy within $\pm2$ bins ($\pm0.08$~Hz) of the reference HR, with the neighborhood restricted to the heartbeat band. It describes how much spectral energy is concentrated near the reference frequency. This quantity varies across positions and has a near-zero linear correlation with $Q_k$ (Pearson $r=-0.03$). Together with the peak-based results in panels (a)--(b), these observations show that favorable motion projection does not guarantee a readable heartbeat peak in each measurement, supporting further assessment from the acquired signal.

\begin{table*}[!t]
\centering
\caption{Reliability-score comparison on simulated and real observations. AUC treats windows with a dominant HR-band peak within $\pm1$ bin of the reference HR as positive. Spear. correlates each method's score with the negative absolute error of HEAR's HR estimate. $Q_k$ is evaluated only in simulation. The shaded row marks HEAR.}
\label{tab:bench}
\setlength{\tabcolsep}{4pt}
\begin{tabularx}{\textwidth}{>{\raggedright\arraybackslash}p{0.115\textwidth}|>{\raggedright\arraybackslash}p{0.165\textwidth}|YY|YY|YY}
\toprule
\multirow{2}{*}{Category} & \multirow{2}{*}{Quality predictor} & \multicolumn{2}{c|}{Simulation\;{\scriptsize$N{=}562{,}986$}} & \multicolumn{2}{c|}{60\,GHz Parralejo\;{\scriptsize$N{=}1408$}} & \multicolumn{2}{c}{120\,GHz VitalSense\;{\scriptsize$N{=}480$}} \\
\cmidrule(lr){3-4}\cmidrule(lr){5-6}\cmidrule(lr){7-8}
 & & AUC\ua & Spear.\ua & AUC\ua & Spear.\ua & AUC\ua & Spear.\ua \\
\midrule
\multirow{1}{*}{Geometric prior} 
 & $Q_k = S\cdot C$ (Eq.~\eqref{eq:qk}) & 0.635 & $+0.165$ & --- & --- & --- & --- \\
\midrule
\multirow{4}{*}{Spectral-shape} & snr\_mean & 0.750 & $+0.284$ & 0.758 & $+0.345$ & 0.852 & $+0.582$ \\
 & energy\_conc & 0.748 & $+0.277$ & 0.747 & $+0.321$ & 0.843 & $+0.577$ \\
 & prominence & 0.742 & $+0.360$ & 0.689 & $+0.279$ & 0.830 & $+0.511$ \\
 & neg\_flatness & 0.669 & $+0.215$ & 0.737 & $+0.349$ & 0.788 & $+0.456$ \\
\midrule
\multirow{4}{*}{Harmonic-aware} & Frequency ratio & 0.606 & $+0.277$ & 0.619 & $+0.110$ & 0.522 & $+0.034$ \\
 & Harmonic distance & 0.557 & $+0.152$ & 0.521 & $+0.047$ & 0.517 & $+0.043$ \\
 & Harmonic order & 0.595 & $+0.254$ & 0.615 & $+0.113$ & 0.512 & $+0.017$ \\
 & Proximity indicator & 0.533 & $+0.117$ & 0.513 & $+0.051$ & 0.502 & $+0.029$ \\
\midrule
\multirow{2}{*}{Learned baseline} & MLP-on-spectrum & 0.868 & $+0.633$ & 0.766 & $+0.380$ & 0.778 & $+0.420$ \\
 & 1D-CNN & 0.890 & $+0.674$ & 0.793 & $+0.413$ & 0.875 & $+0.598$ \\
\midrule
\rowcolor{oursgray}
Ours & HEAR gate & 0.868 & $+0.655$ & 0.856 & $+0.572$ & 0.941 & $+0.736$ \\
\bottomrule
\end{tabularx}
\end{table*}

\subsection{Zero-Shot Observability Assessment}
\label{sec:transfer}

Under the zero-shot protocol in Sec.~\ref{sec:setup}, we evaluate whether the simulation-trained HEAR gate assesses heartbeat observability on real radar measurements without real-data adaptation. As shown in Table~\ref{tab:bench}, HEAR achieves an AUC of 0.868 in simulation, 0.856 on Parralejo 60~GHz, and 0.941 on VitalSense 120~GHz.

On the two real datasets, the strongest hand-crafted indicator achieves AUCs of 0.758 and 0.852, while 1D-CNN reaches 0.793 and 0.875. Although 1D-CNN has a higher simulation AUC than HEAR (0.890 versus 0.868), its AUC is lower on both real datasets. Thus, higher simulation discrimination does not necessarily translate into better performance on real measurements. HEAR also yields positive correlations between its score and the negative absolute HR-head error, with Spearman coefficients of 0.572 and 0.736 on Parralejo and VitalSense, respectively. These results support the transfer of the learned observability score to the two evaluated real datasets. Performance differences across acquisition conditions are examined in Sec.~\ref{sec:diag}.

\begin{figure*}[t]
\centering
\includegraphics[width=\textwidth]{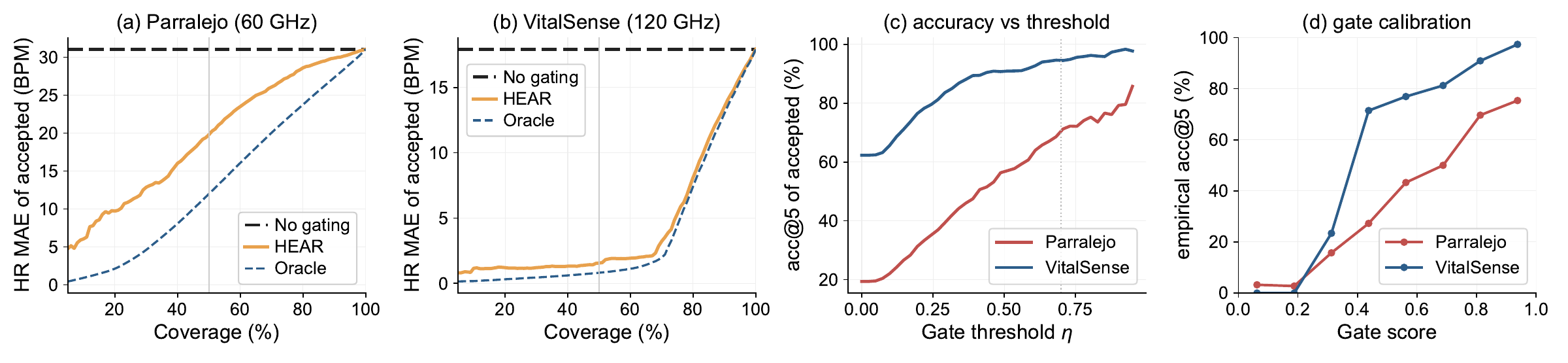}
\caption{Observability-guided selective HR estimation on real sliding windows.
All HR estimates come from HEAR's HR head.
(a)--(b) Risk--coverage curves on Parralejo 60~GHz and VitalSense 120~GHz, comparing selection by HEAR scores with Oracle selection by reference-based absolute errors. The horizontal dashed line marks MAE at full coverage.
(c) Accepted-window Acc@5 BPM versus the selection threshold $\eta$.
(d) Empirical Acc@5 BPM within observability-score bins on both datasets.}
\label{fig:riskcov}
\end{figure*}

\begin{table*}[!t]
\centering
\caption{HR estimation performance across 10 equal-width observability-score bins. Sample count $n$, MAE (BPM), and accuracy are computed within each bin. Coverage (cov) is the fraction of all windows with scores at or above the bin's lower bound. Simulation uses FFT-peak estimates with accuracy measured within $\pm1$ bin ($\pm2.4$~BPM). Real datasets use HEAR's HR-head estimates with accuracy measured within $\pm5$~BPM.}
\label{tab:simcal}
\setlength{\tabcolsep}{2pt}\footnotesize
\begin{tabularx}{\textwidth}{>{\raggedright\arraybackslash}p{0.085\textwidth}|YYYY|YYYY|YYYY}
\toprule
\multirow{2}{*}{Gate bin} & \multicolumn{4}{c|}{Simulation\;{\scriptsize$N{=}562{,}986$}} & \multicolumn{4}{c|}{60\,GHz Parralejo\;{\scriptsize$N{=}1408$}} & \multicolumn{4}{c}{120\,GHz VitalSense\;{\scriptsize$N{=}480$}} \\
\cmidrule(lr){2-5}\cmidrule(lr){6-9}\cmidrule(lr){10-13}
 & $n$ & cov & MAE\da & acc\ua & $n$ & cov & MAE\da & acc\ua & $n$ & cov & MAE\da & acc\ua \\
\midrule
$[0.0,0.1)$ & 108{,}718 & 100\% & 44.9 & 5.6\% & 218 & 100\% & 41.3 & 4\% & 25 & 100\% & 58.2 & 0\% \\
$[0.1,0.2)$ & 177{,}403 & 80.7\% & 34.6 & 9.6\% & 375 & 84.5\% & 43.0 & 2\% & 66 & 94.8\% & 57.1 & 0\% \\
$[0.2,0.3)$ & 63{,}757 & 49.2\% & 25.6 & 18.6\% & 226 & 57.9\% & 38.8 & 8\% & 38 & 81.0\% & 49.5 & 8\% \\
$[0.3,0.4)$ & 38{,}523 & 37.9\% & 20.1 & 28.5\% & 172 & 41.8\% & 26.3 & 21\% & 37 & 73.1\% & 23.0 & 41\% \\
$[0.4,0.5)$ & 29{,}437 & 31.0\% & 15.5 & 38.5\% & 120 & 29.6\% & 19.7 & 28\% & 24 & 65.4\% & 3.7 & 75\% \\
$[0.5,0.6)$ & 24{,}927 & 25.8\% & 11.6 & 49.8\% & 79 & 21.1\% & 11.8 & 43\% & 41 & 60.4\% & 2.8 & 81\% \\
$[0.6,0.7)$ & 22{,}827 & 21.4\% & 8.5 & 60.4\% & 69 & 15.5\% & 14.8 & 49\% & 30 & 51.9\% & 4.6 & 77\% \\
$[0.7,0.8)$ & 22{,}013 & 17.3\% & 5.8 & 71.4\% & 51 & 10.6\% & 9.5 & 55\% & 32 & 45.6\% & 2.1 & 84\% \\
$[0.8,0.9)$ & 24{,}733 & 13.4\% & 3.4 & 82.7\% & 43 & 7.0\% & 5.3 & 72\% & 47 & 39.0\% & 1.6 & 94\% \\
\rowcolor{oursgray}
$[0.9,1.0)$ & 50{,}648 & 9.0\% & 0.86 & 95.3\% & 55 & 3.9\% & 4.4 & 78\% & 140 & 29.2\% & 1.2 & 97\% \\
\bottomrule
\end{tabularx}
\end{table*}

\subsection{Observability-Guided Selective Estimation}
\label{sec:fusion}

We first examine the relationship between the gate score and HR-estimation performance by dividing windows into 10 equal-width score bins (Table~\ref{tab:simcal}). Higher-score groups generally have lower MAE and higher accuracy. On VitalSense, the highest-score group, $g\in[0.9,1.0)$, contains 140 windows, or 29.2\% of the dataset, and achieves 1.2~BPM MAE and 97\% Acc@5 BPM. These trends support using the score to rank observations by estimation reliability.

We then evaluate selective estimation under the fixed-coverage protocol in Sec.~\ref{sec:setup}. Fig.~\ref{fig:riskcov} and Table~\ref{tab:riskcov} report the risk--coverage results. Retaining the highest-scoring 50\% of windows reduces VitalSense MAE from 17.9 to 1.6~BPM and improves Acc@5 BPM from 62\% to 93\%. At the same coverage, Parralejo MAE decreases from 31.0 to 19.8~BPM. Selection improves estimation on both real datasets, although the retained-window error remains higher on Parralejo.

At every reported coverage level on the two real datasets, HEAR's MAE is no higher than that obtained using the SNR baseline. HEAR also achieves lower AURC than this baseline in all three datasets, with 11.2 versus 13.6 in simulation, 19.8 versus 24.4 on Parralejo, and 4.4 versus 8.0 on VitalSense. These comparisons show that the learned score provides more effective selective estimation than the evaluated SNR criterion over the reported coverage range.

\begin{table*}[!t]
\centering
\caption{Risk--coverage trade-off of selective HR estimation. All selectors use the same HEAR HR-head estimates. Entries report retained-window MAE (BPM) / Acc@5 BPM (\%). Full coverage denotes no selection. HEAR and snr\_m select windows by observability and SNR scores, respectively. Oracle selects windows with the smallest reference-based absolute errors. The final row reports normalized AURC over 5\%--100\% coverage. Lower is better.}
\label{tab:riskcov}
\setlength{\tabcolsep}{2pt}
\begin{tabularx}{\textwidth}{l|YYY|YYY|YYY}
\toprule
\multirow{2}{*}{Coverage} & \multicolumn{3}{c|}{Simulation\;{\scriptsize$N{=}562{,}986$}} & \multicolumn{3}{c|}{60\,GHz Parralejo\;{\scriptsize$N{=}1408$}} & \multicolumn{3}{c}{120\,GHz VitalSense\;{\scriptsize$N{=}480$}} \\
\cmidrule(lr){2-4}\cmidrule(lr){5-7}\cmidrule(lr){8-10}
 & HEAR & snr\_m & Oracle & HEAR & snr\_m & Oracle & HEAR & snr\_m & Oracle \\
\midrule
100\% (no gating) & 16.6\,/\,53 & 16.6\,/\,53 & 16.6\,/\,53 & 31.0\,/\,19 & 31.0\,/\,19 & 31.0\,/\,19 & 17.9\,/\,62 & 17.9\,/\,62 & 17.9\,/\,62 \\
90\% & 16.7\,/\,53 & 16.4\,/\,54 & 11.3\,/\,59 & 29.9\,/\,21 & 29.9\,/\,21 & 27.3\,/\,22 & 13.5\,/\,69 & 15.1\,/\,68 & 13.0\,/\,69 \\
70\% & 15.8\,/\,55 & 15.6\,/\,56 & 3.9\,/\,75 & 26.1\,/\,27 & 27.4\,/\,26 & 20.0\,/\,28 & 3.2\,/\,86 & 9.8\,/\,80 & 2.0\,/\,89 \\
60\% & 14.8\,/\,57 & 15.1\,/\,58 & 2.0\,/\,88 & 23.5\,/\,31 & 26.4\,/\,28 & 16.1\,/\,32 & 1.9\,/\,91 & 8.2\,/\,84 & 1.1\,/\,100 \\
50\% & 13.1\,/\,61 & 14.3\,/\,60 & 1.0\,/\,100 & 19.8\,/\,36 & 25.1\,/\,30 & 12.0\,/\,39 & 1.6\,/\,93 & 8.6\,/\,85 & 0.8\,/\,100 \\
40\% & 10.3\,/\,67 & 13.3\,/\,64 & 0.3\,/\,100 & 16.0\,/\,42 & 23.5\,/\,34 & 8.1\,/\,48 & 1.3\,/\,96 & 5.8\,/\,90 & 0.6\,/\,100 \\
30\% & 6.4\,/\,76 & 11.9\,/\,68 & 0.0\,/\,100 & 12.9\,/\,48 & 21.9\,/\,38 & 4.6\,/\,65 & 1.2\,/\,97 & 4.3\,/\,93 & 0.5\,/\,100 \\
\midrule
\rowcolor{oursgray}
AURC$\downarrow$ & 11.2 & 13.6 & 3.5 & 19.8 & 24.4 & 13.7 & 4.4 & 8.0 & 3.6 \\
\bottomrule
\end{tabularx}
\end{table*}

\begin{table*}[!t]
\centering
\caption{Selective HR estimation with a shared HEAR score. ``All'' denotes all windows, and ``After gate'' denotes windows accepted by $g\geq0.7$, retaining approximately 17\%, 11\%, and 46\% of windows in simulation, Parralejo, and VitalSense, respectively. Within each dataset, all estimators use the same retained windows. $\Delta$ reports the gain in percentage points. Accuracy is measured within $\pm1$ bin in simulation and $\pm5$~BPM on real data. Dashes indicate unreported results.}
\label{tab:methodcmp}
\setlength{\tabcolsep}{4pt}
\begin{tabularx}{\textwidth}{l|YYY|YYY|YYY}
\toprule
& \multicolumn{3}{c|}{Simulation (acc@$\pm1$\,bin)} & \multicolumn{3}{c|}{Parralejo 60\,GHz} & \multicolumn{3}{c}{VitalSense 120\,GHz} \\
\cmidrule(lr){2-4}\cmidrule(lr){5-7}\cmidrule(lr){8-10}
HR estimator & All & After gate\ua & $\Delta$ & All & After gate\ua & $\Delta$ & All & After gate\ua & $\Delta$ \\
\midrule
FFT peak & 29.8\% & 86.7\% & $+56.9$ & 32.0\% & 65.8\% & $+33.8$ & 81.7\% & 94.1\% & $+12.4$ \\
Weighted peak & 30.0\% & 86.8\% & $+56.9$ & 32.7\% & 66.4\% & $+33.7$ & 83.1\% & 93.6\% & $+10.5$ \\
Autocorrelation~\cite{Seifizarei2025} & --- & --- & --- & 36.2\% & 59.7\% & $+23.6$ & 83.3\% & 90.0\% & $+6.6$ \\
Harmonic product spectrum & 22.4\% & 42.6\% & $+20.2$ & 19.7\% & 37.6\% & $+17.8$ & 69.0\% & 77.6\% & $+8.7$ \\
VMD~\cite{Yang2025} & --- & --- & --- & 32.7\% & 65.1\% & $+32.4$ & 76.5\% & 90.9\% & $+14.4$ \\
\rowcolor{oursgray}
Ours (network HR head) & 45.9\% & 87.1\% & $+41.2$ & 19.4\% & 68.5\% & $+49.1$ & 62.3\% & 94.5\% & $+32.2$ \\
\bottomrule
\end{tabularx}
\end{table*}

\begin{table*}[!t]
\centering
\caption{Input feature ablation for the 2-D token $[X(b),r(b)]$. AUC measures observability discrimination. SE denotes bootstrap standard error, and $\Delta$AUC is the ablated model's AUC minus the full-token model's AUC. The ratio-only variant performs poorly on VitalSense, where numerical results are not reported. Dashes mark omitted baseline differences and unreported results.}
\label{tab:ablation}
\setlength{\tabcolsep}{2.5pt}
\begin{tabularx}{\textwidth}{l|YY|YYY|YYY}
\toprule
 & \multicolumn{2}{c|}{Simulation test set} & \multicolumn{3}{c|}{60\,GHz Parralejo} & \multicolumn{3}{c}{120\,GHz VitalSense} \\
 & \multicolumn{2}{c|}{\scriptsize $N{=}562{,}986$} & \multicolumn{3}{c|}{\scriptsize $N{=}1408$} & \multicolumn{3}{c}{\scriptsize $N{=}480$} \\
\cmidrule(lr){2-3}\cmidrule(lr){4-6}\cmidrule(lr){7-9}
Configuration & AUC\ua & $\Delta$AUC & AUC\ua & SE & $\Delta$AUC & AUC\ua & SE & $\Delta$AUC \\
\midrule
\rowcolor{oursgray}
2-D token $[X(b),\, f/f_{\mathrm{b}}]$ (full) & 0.865 & --- & 0.864 & 0.013 & --- & 0.941 & 0.011 & --- \\
\midrule
\quad $-$ spectral value $X(b)$ & 0.517 & $-0.348$ & 0.325 & 0.052 & $-0.539$ & --- & --- & --- \\
\quad $-$ frequency-ratio encoding & 0.784 & $-0.081$ & 0.627 & 0.049 & $-0.237$ & 0.832 & 0.027 & $-0.109$ \\
\bottomrule
\end{tabularx}
\end{table*}

\subsection{Shared Quality Gating Across Heart-Rate Estimators}\label{sec:methodcmp}

We test whether the same HEAR score can guide measurement selection for different HR estimators. Using the fixed threshold $g\geq0.7$, all estimators within each dataset receive the same retained windows. This selection retains approximately 17\% of simulation windows, 11\% of Parralejo windows, and 46\% of VitalSense windows (Table~\ref{tab:methodcmp}).

Every evaluated external estimator achieves higher accuracy on the retained windows than on the full set, although the gains vary across methods and datasets. FFT peak accuracy increases from 32.0\% to 65.8\% on Parralejo and from 81.7\% to 94.1\% on VitalSense. VMD accuracy increases from 32.7\% to 65.1\% and from 76.5\% to 90.9\%, respectively. The harmonic product spectrum also benefits from selection, but its retained-window accuracy remains lower than that of the other evaluated estimators.

HEAR's own HR head also benefits from the same selection and achieves the highest reported accuracy on the retained windows in each dataset, with 87.1\% Acc@$\pm1$ bin in simulation, 68.5\% Acc@5 BPM on Parralejo, and 94.5\% Acc@5 BPM on VitalSense. Together, these results show that a single learned score can identify observations that support more accurate estimation across the evaluated HR estimators.

\subsection{Input Feature Ablation}\label{sec:ablation}

We evaluate the contribution of each component of the 2-D token $[X(b),r(b)]$ by removing one component at a time and retraining with the same recipe. Table~\ref{tab:ablation} reports gate-discrimination AUC on the simulation test set and the two real datasets. The full-token model achieves AUCs of 0.865, 0.864, and 0.941, respectively. These values differ slightly from Table~\ref{tab:bench} because the ablation models are independently retrained.

Removing the spectral magnitude $X(b)$ reduces AUC to 0.517 in simulation and 0.325 on Parralejo, showing that the frequency ratio alone does not reliably assess heartbeat readability in these settings. This variant also performs poorly on VitalSense, for which quantitative results are not reported.

Removing the frequency ratio $r(b)$ reduces AUC to 0.784 in simulation, 0.627 on Parralejo, and 0.832 on VitalSense. The ratio therefore improves observability discrimination in the evaluated model. Because $r(b)$ supplies both a frequency coordinate and a relation to the respiration fundamental, this ablation measures their combined contribution. Among the reported configurations, jointly using spectral magnitude and frequency ratio achieves the highest AUC on all three datasets.

\subsection{Stratified Performance Analysis}\label{sec:diag}

Differences in acquisition conditions and participant populations prevent attributing the overall performance contrast between the two real datasets to carrier frequency alone. To examine variation within a dataset, we stratify the full Parralejo sliding-window set ($N=1408$) by posture, physiological state, and heart-rate range (Table~\ref{tab:strat}). AUC is 0.897 for lying and 0.771 for sitting, and 0.914 at rest and 0.759 after exercise. The joint Lying$\times$Rest subset achieves an AUC of 0.932. These results show that discrimination varies across measurement conditions and is lower in sitting and post-exercise observations.

In the $>110$~BPM subset, which contains 54 windows, AUC falls to 0.327, indicating poor discrimination on this subset. Its small size limits the conclusions that can be drawn about high-HR performance more broadly. This subset also includes reference heart rates both within and outside the deployed $[0.8,2.0)$~Hz band. Its aggregate AUC therefore does not isolate performance on high-HR observations within the output band. The implications for high-HR sensing are discussed in Sec.~\ref{sec:discussion}.

\begin{figure*}[t]
\centering
\includegraphics[width=0.9\textwidth]{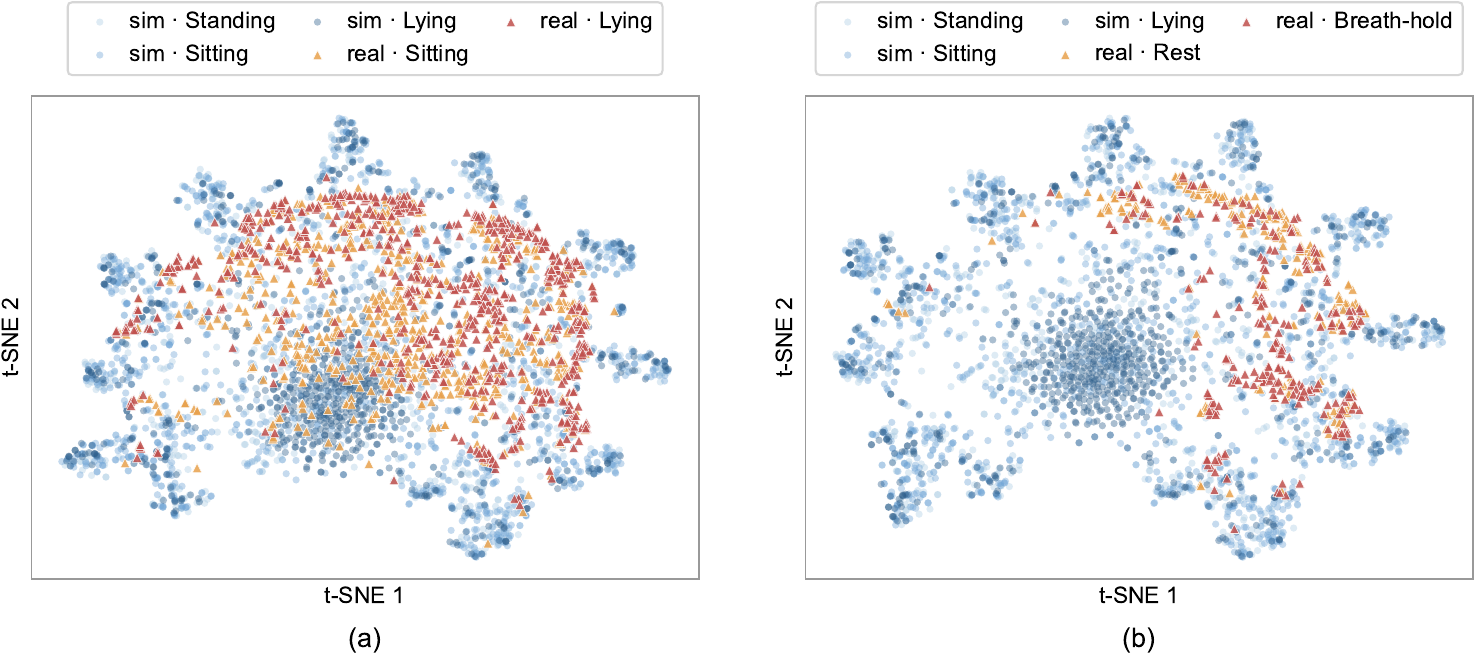}
\caption{Joint t-SNE visualization of 75-dimensional normalized phase spectra from simulation and real datasets.
(a) Simulation versus Parralejo 60~GHz, colored by posture.
(b) Simulation versus VitalSense 120~GHz, with simulated observations colored by posture and real observations by recording condition.
Circles and triangles represent simulated and real observations, respectively. The two-dimensional embedding shows partial overlap between simulated and real observations together with dataset-specific variation.}
\label{fig:tsne}
\end{figure*}

Fig.~\ref{fig:tsne} provides a complementary t-SNE visualization of the 75-dimensional phase spectra. The two-dimensional embedding shows overlap between simulated and real observations together with dataset-specific variation, providing a qualitative view of the feature distributions.

\begin{table}[!t]
\centering
\caption{Stratified AUC of the HEAR observability score on the full Parralejo sliding-window set ($N=1408$). Windows are grouped by reference HR range, posture, physiological state, and the joint Lying$\times$Rest subset. $n$ denotes the number of windows. The $>110$~BPM subset includes reference HRs both within and outside the deployed HR band.}
\label{tab:strat}
\begin{tabular*}{\columnwidth}{@{\extracolsep{\fill}}llcc}
\toprule
Dimension & Subset & $n$ & AUC\ua \\
\midrule
\multirow{4}{*}{HR range}
& $<70$~BPM & 661 & 0.918 \\
& 70--90~BPM & 533 & 0.877 \\
& 90--110~BPM & 160 & 0.843 \\
& $>110$~BPM & 54 & 0.327 \\
\midrule
\multirow{2}{*}{Posture}
& Lying & 706 & 0.897 \\
& Sitting & 702 & 0.771 \\
\midrule
\multirow{2}{*}{State}
& Rest & 704 & 0.914 \\
& Post-exercise & 704 & 0.759 \\
\midrule
Joint & Lying $\times$ Rest & 353 & 0.932 \\
\bottomrule
\end{tabular*}
\end{table}

\subsection{Computational Efficiency and Edge Deployment}\label{sec:efficiency}

We benchmark HEAR on an NVIDIA Jetson AGX Orin in 30~W mode using only the ARMv8 CPU, without GPU or DLA acceleration (Table~\ref{tab:efficiency}). We use HEAR's own HR head so that a single forward pass provides both the observability score and HR estimate. Processing a 25~s phase-signal window takes 50.8~ms, including 47.7~ms for signal preprocessing and 3.1~ms for network inference.

The network contains 5{,}295 parameters and requires 0.26~M FLOPs per window. Its FP32 weights occupy approximately 20.7~KB, while the peak memory usage of the full runtime process is approximately 1.0~GB. Signal preprocessing accounts for most of the processing time. The total processing time is well below the 10~s window hop used in the real-data evaluation, supporting online updates as successive observation windows become available.

\begin{table}[!t]
\centering
\caption{Computational efficiency of HEAR on an NVIDIA Jetson AGX Orin in 30~W mode using only the ARMv8 CPU. Times are reported per 25~s phase-signal window. Total processing time includes signal preprocessing and network inference. A single forward pass produces both outputs. Peak memory refers to the full runtime process.}
\label{tab:efficiency}
\setlength{\tabcolsep}{4pt}
\begin{tabular*}{\columnwidth}{@{\extracolsep{\fill}}ll}
\toprule
Metric & Value \\
\midrule
Network parameters & 5{,}295 \\
Model weights (FP32) & 20.7~KB \\
Network FLOPs / window & 0.26~M \\
Signal preprocessing (per window) & 47.7~ms \\
Network inference (per window) & 3.1~ms \\
Total processing time (per window) & 50.8~ms \\
Peak process memory & $\sim$1.0~GB \\
\bottomrule
\end{tabular*}
\end{table}

\section{Discussion and Limitations}\label{sec:discussion}

HEAR learns to assess heartbeat readability from acquired measurements and uses the resulting score to select observations for HR estimation. The improved accuracy on retained windows comes at the cost of reduced coverage. Rejected windows provide no HR output under this selection rule. The results therefore support applications that accept intermittent reliable readings, while continuous monitoring requires further evaluation of the duration and frequency of gaps in accepted outputs.

The observability label measures agreement between the dominant heartbeat-band peak and the reference heart rate. It provides an operational measure of spectral readability, but a heartbeat component that is not dominant may still be recoverable by an estimator using additional spectral structure. The cross-estimator results in Sec.~\ref{sec:methodcmp} support the score's utility for the evaluated methods. Its usefulness for other estimators remains to be established.

The stratified results in Sec.~\ref{sec:diag} show weaker discrimination in high-HR and post-exercise observations, with stronger evidence for performance under resting conditions. The $>110$~BPM subset contains only 54 windows and includes reference rates both within and outside the deployed HR band, limiting conclusions about the cause and extent of this failure. The simulator also uses simplified scatterer amplitudes, prescribed chest-wall motion, fixed spatial motion weights, and a fixed radar signal chain. These assumptions enable controlled analysis and training but provide limited coverage of clothing variation, body variability, pose transitions, environmental multipath, and hardware diversity.

Future work should include broader high-HR evaluation that distinguishes reference rates within and outside the output band, together with more diverse body, motion, and radar configurations. Favorable observation geometry and measurement-level assessment can also be combined in active sensing, using the learned score to guide radar position adjustments. Adaptive selection thresholds could further accommodate application-specific requirements for accuracy and coverage.

\section{Conclusion}\label{sec:expand}

This paper studied learning to assess heartbeat readability from acquired phase spectra for selective mmWave heart-rate estimation. Our physical analysis shows that coherent superposition of scatterer returns can alter the local heartbeat phase response under fixed motion projections, motivating assessment beyond macroscopic geometry. We developed a controllable FMCW simulator to derive observation-level readability labels and proposed HEAR, a compact dual-task Transformer that jointly predicts an observability score and heart rate. Trained solely on simulated observations, HEAR transfers zero-shot to two public real datasets collected at 60 and 120~GHz. Its score improves selective estimation with both its own HR head and the evaluated external estimators. On the 120~GHz dataset, selection reduces HR-head MAE from 17.9 to 1.6~BPM at 50\% coverage. Signal preprocessing and network inference together take 50.8~ms per window on an edge device, supporting online use of the learned assessment.

\bibliographystyle{IEEEtran}
\bibliography{BIB/refs}

\clearpage
\appendices

\section{Implementation Details}\label{app:impl}

\begin{table}[!t]
\centering
\caption{Implementation details and hyper-parameters. The table summarizes the signal-processing settings, network architecture, training configuration, and simulation parameters used for reproducibility.}
\label{tab:impl}
\setlength{\tabcolsep}{3pt}
\begin{tabular*}{\columnwidth}{@{\extracolsep{\fill}}l >{\raggedright\arraybackslash}p{0.36\columnwidth} >{\raggedright\arraybackslash}p{0.42\columnwidth}}
\toprule
Category & Parameter & Value \\
\midrule
\multirow{3}{*}{Signal proc.} & Frame rate / analysis window & 20~Hz / 500 frames (25~s)\\
 & Spectrum crop & $[0,3.0)$~Hz (75 bins), $L^2$-norm.\ after crop\\
 & Heartbeat / respiration band & $[0.8,2.0)$~Hz (30 bins) / $[0.1,0.6]$~Hz ($f_{\mathrm{b}}$: band argmax) \\
\midrule
\multirow{2}{*}{Architecture} & token / embed dim & 2 / 16 \\
 & attention / layers / params & std.\ 4-head / 2 / 5{,}295 \\
\midrule
\multirow{3}{*}{Training} & Optimizer & Adam (lr $10^{-3}$, wd $10^{-4}$), cosine annealing\\
 & batch / epochs & 256 / 30 (joint heads)\\
 & test split & 10\% held-out body-drive pairs (seed 42)\\
\midrule
\multirow{2}{*}{Simulation} & bodies / drives & 48 (12 post.$\times$4 types) / 2240 distinct (107{,}520 instances) \\
 & carrier / BW / nominal SNR / range & 58.48~GHz / 5.86~GHz / 50~dB / 0.40~m offset \\
\bottomrule
\end{tabular*}
\end{table}

Table~\ref{tab:impl} summarizes the signal-processing parameters, network architecture, training settings, and simulation hyper-parameters.

The physiological drive library that animates the simulator is extracted from mmWave radar recordings collected by the authors on a separate acquisition setup. These source recordings share no subject, device, or recording with the Parralejo or VitalSense evaluation datasets. {Neither evaluation dataset is used for training, validation, model selection, or threshold tuning. The self-collected recordings provide the respiratory and heartbeat drive waveforms that animate the simulator, and all training observations seen by the network are generated by simulation.}

The source recordings were acquired with a custom development board based on an Infineon 60-GHz mmWave radar chip with one transmit and three receive antennas, using the receive channel with the highest SNR. The radar operates at a carrier frequency of 58.48~GHz with 5.86~GHz sweep bandwidth and a 20~Hz frame rate, matching the FMCW parameters used in simulation. Recordings were collected from two healthy subjects seated at a subject--radar distance of approximately 0.5~m. {We divided the recordings into non-overlapping 25~s segments. All recordings were acquired with the subject seated and stationary. Each segment was then screened by the ratio of its heart-band (0.8--2.0~Hz) spectral peak to the heart-band median magnitude, and segments with a ratio of at least 4 were retained as having clear heartbeat content. This procedure yielded 40 source segments for drive extraction.}

{During library augmentation, respiration drives are frequency-shifted with the same analytic-signal method as heartbeat drives, with target frequencies drawn uniformly at random from the 0.04~Hz grid over $[0.1,0.5]$~Hz. Each frequency-shifted drive preserves the amplitude scale of its source waveform, and amplitudes are additionally jittered by uniform random factors in $[0.7,1.3]$ for heartbeat drives and $[0.8,1.2]$ for respiration drives. In the FMCW simulation, drive waveforms are converted from phase to displacement by the $\lambda/(4\pi)$ factor of the source radar. Complex Gaussian noise $n\sim\mathcal{CN}(0,\sigma_n^2)$ with $\sigma_n=10^{-\mathrm{SNR}/20}$ at a nominal $\mathrm{SNR}=50$~dB is added to the per-scatterer-averaged mixed signal, where the reference is a unit-amplitude scatterer return; since the averaged sum has magnitude at most one, this is a fixed nominal noise setting rather than the realized SNR of each simulated observation.}

\section{Public Datasets, Preprocessing, and Windowing}
\label{app:dataset}

Parralejo~\cite{Parralejo2026} is a 60~GHz FMCW radar dataset with carrier frequency 60.25~GHz, 480~MHz sweep bandwidth, 64 ADC samples, 32 chirps per frame, and 10~Hz frame rate. It contains 110 subjects, including healthy and cardiac-condition participants with diverse age and BMI. Each subject was recorded in Lying and Sitting postures under Rest and Post-exercise conditions, yielding $110\times2\times2=440$ recordings. The dataset provides range-FFT sequences and synchronized ECG ground truth.

VitalSense~\cite{Wu2026_VitalSense} is a 120~GHz dataset with carrier frequency 122--123~GHz. It contains 24 subjects recorded under Rest and Breath-hold scenarios, yielding 48 recordings. The dataset provides pre-extracted vital-sign displacement signals at approximately 333~Hz and per-second HR ground truth from a Mindray ePM10 monitor.

For Parralejo, we average chirps within each frame to obtain a frame-level range FFT, select the maximum-energy range bin within 0.3--1.5~m, extract the corresponding slow-time complex sequence, and unwrap its phase. For VitalSense, processing starts from the provided displacement sequence. Each record is segmented into 25~s sliding windows with a 10~s hop. Each window is linearly resampled to 500 samples at 20~Hz and processed by the same preprocessing chain used for simulation training, including inter-frame differencing, Hampel filtering with window size 11 and threshold $3\sigma$, and 5-point moving-average smoothing. {The magnitude spectrum of each processed window is cropped to $[0,3.0)$~Hz (75 bins on the 0.04~Hz grid) and $L^2$-normalized within this range, identically to the simulated spectra, before being fed to HEAR.}

For Parralejo, the ECG reference is band-pass filtered at 5--30~Hz, and R peaks are detected from the absolute-valued filtered signal. The reference HR of each window is computed from the median inter-beat interval within the same window. For VitalSense, the reference HR is the median per-second Mindray HR within each window.

After windowing, Parralejo yields 1{,}408 windows and VitalSense yields 480 windows. The true-HR distributions are dominated by resting rates, with median HRs of 71 and 65~BPM for Parralejo and VitalSense, respectively. Parralejo contains a small number of high-HR windows above 120~BPM, accounting for 1.7\% of windows with a maximum of 144~BPM, mainly from post-exercise recordings. All VitalSense windows are $\leq83$~BPM. The per-scenario HR distributions are shown in Fig.~\ref{fig:hr_dist}.
\section{Sliding-Window Hop Sweep}\label{app:hopsweep}

\begin{table}[!t]
\centering
\caption{Sliding-window hop sensitivity on all 440 Parralejo records with a fixed 25~s window and $\eta=0.7$. Smaller hops raise temporal coverage while gate-discrimination AUC stays stable for 2 to 15~s hops. Temporal cov. is the per-record fraction of time covered by the union of accepted 25~s windows, averaged over records, and Rec. is the fraction of records with at least one accepted window.}
\label{tab:hopsweep}
\setlength{\tabcolsep}{4pt}\footnotesize
\begin{tabular*}{\columnwidth}{@{\extracolsep{\fill}}cccccc@{}}
\toprule
Hop\,(s) & Windows & AUC\ua & Temporal cov.\ua & Acc@5 BPM\ua & Rec.\ua \\
\midrule
2 & 6159 & 0.871 & 23\% & 70\% & 36\% \\
5 & 2815 & 0.863 & 20\% & 69\% & 33\% \\
10 & 1408 & 0.856 & 15\% & 68\% & 24\% \\
15 & 970 & 0.864 & 15\% & 73\% & 23\% \\
25 & 877 & 0.836 & 9\% & 60\% & 16\% \\
\bottomrule
\end{tabular*}
\end{table}

Table~\ref{tab:hopsweep} reports the sliding-window hop sweep on Parralejo. A 5~s hop can improve temporal coverage with moderate additional compute, whereas the 10~s hop is used for evaluation to reduce correlation between adjacent windows.

\end{document}